\documentclass[11pt]{article}

\usepackage[T1]{fontenc}

\usepackage[colorlinks=true, allcolors=blue,citecolor=blue]{hyperref}
\usepackage{abstract, adjustbox, amsmath, amsthm, amssymb, bm, booktabs, caption, colortbl, etoolbox, lettrine, lipsum, listofitems, lmodern, fancyhdr, graphicx, mathtools, multirow, subfigure, tabularx, titlesec, tikz, titling, wrapfig, xcolor, xurl}
\usepackage[subfigure]{tocloft}

\usepackage{ebgaramond}
\usepackage{CJKutf8}

\newcommand{\dropcap}[1]{\lettrine[lines=2,lraise=0.05,findent=0.1em, nindent=0em]{{\normalfont{#1}}}{}}
\usepackage[english]{babel}

\usepackage[margin=0.8in, top=1.2in, bottom=1in]{geometry}

\pretitle{
    \begin{flushleft}
    \vspace{-15mm} 
    \Huge
    \bfseries 
} 
\posttitle{
    \end{flushleft}
    \hrule
    \vspace{3mm}
} 
\preauthor{\begin{flushleft}\large}
\postauthor{\end{flushleft}}

\definecolor{c_yellow}{HTML}{F5DDA9}
\definecolor{c_orange}{HTML}{EC9F7E}
\definecolor{c_red}{HTML}{EC7070}
\definecolor{c_pink}{HTML}{F8A6A6}
\definecolor{c_purple}{HTML}{A07CB0}
\definecolor{c_lightblue}{HTML}{97c3d0}
\definecolor{c_darkblue}{HTML}{2F7194}
\definecolor{c_lightgreen}{HTML}{AFD8BC}
\definecolor{c_darkgreen}{HTML}{48675A}
\definecolor{c_lightbrown}{HTML}{C6BFA2}
\definecolor{c_lightgrey}{HTML}{AFC1B9}
\definecolor{c_darkgrey}{HTML}{3D4244}

\definecolor{lightblue}{HTML}{ECF3F6}
\definecolor{darkblue}{HTML}{D0E6F0}
\definecolor{linkcolor}{HTML}{1B6EA2}

\hypersetup{colorlinks=true,linkcolor=linkcolor,urlcolor=linkcolor,citecolor=linkcolor,allcolors=linkcolor}
\renewenvironment{abstract}
 {\par\noindent\textbf{\abstractname}\ \ignorespaces}
 {\par\medskip}

\renewcommand\thesection{\Roman{section}} 
\renewcommand\thesubsection{\roman{subsection}}

\titleformat{\section}[block]{\Large\bfseries}
{\thesection.}{1em}{} 
\titleformat{\subsection}[block]{\large\bfseries}{\thesubsection.}{1em}{} 

\adjustboxset{margin=10pt,bgcolor=lightblue,rndcorners=6}
\DeclareCaptionFormat{pnasformat}{\fontsize{9}{11}\selectfont#1#2#3}
\DeclareRobustCommand{\tablefont}{%
        \fontencoding{\encodingdefault}%
        \fontseries{m}
        \fontshape{n}
        \fontsize{9}{11}
        \selectfont}
\DeclareTextFontCommand{\texttable}{\tablefont}

\date{}

\usepackage{enumitem}
\setitemize{leftmargin=5.5mm, rightmargin=3mm}

\title{Deep learning four decades of human migration} 

\author{
	\bfseries{Thomas Gaskin$^{\text{1},\text{2},\star}$}, \bfseries{Guy J. Abel$^{\text{3,4}}$}
}

\begin{document}

\newcommand{\cs}[1]{\fontsize{9}{12}\selectfont{#1}}
\captionsetup{labelfont=bf}
\maketitle
\vspace{-20mm}
{\normalsize\flushleft
\textsuperscript{\textbf{1}}~Department of Applied Mathematics and Theoretical Physics, University of Cambridge, Cambridge CB3 0WA, United Kingdom; \textsuperscript{\textbf{2}}~Department of Mathematics, Imperial College London, London SW7 2AZ, United Kingdom; \textsuperscript{\textbf{3}}~Department of Sociology, Faculty of Social Sciences, University of Hong Kong, Hong Kong SAR, China; \textsuperscript{\textbf{4}}~International Institute for Applied Systems Analysis, Laxenburg 2361, Austria

\medskip $^\star$Correspondence: trg34@cam.ac.uk
}
\vspace{6mm}
\hfill \hrule

\vspace{6mm}

\begin{abstract} 
\noindent We present a novel and detailed dataset on origin-destination annual migration flows and stocks between 230 countries and regions, spanning the period from 1990 to the present. Our flow estimates are further disaggregated by country of birth, providing a comprehensive picture of migration over the last 35 years. The estimates are obtained by training a deep recurrent neural network to learn flow patterns from 18 covariates for all countries, including geographic, economic, cultural, societal, and political information. The recurrent architecture of the neural network means that the entire past can influence current migration patterns, allowing us to learn long-range temporal correlations. By training an ensemble of neural networks and additionally pushing uncertainty on the covariates through the trained network, we obtain confidence bounds for all our estimates, allowing researchers to pinpoint the geographic regions most in need of additional data collection. We validate our approach on various test sets of unseen data, demonstrating that it significantly outperforms traditional methods estimating five-year flows while delivering a significant increase in temporal resolution. The model is fully open source: all training data, neural network weights, and training code are made public alongside the migration estimates, providing a valuable resource for future studies of human migration.

\bigskip

\end{abstract}
\vspace{6mm}
\hrule
\vspace{1mm}
\setcounter{tocdepth}{1}
\tableofcontents
\vspace{4mm}

\newpage
\twocolumn
\hrule height 0.1cm 

\section{Introduction}

\dropcap{T}he movement of people---both within countries and between them---is an important topic across multiple domains. Migration drives demographic change, shaping the size and composition of populations; it can influence labour markets \cite{Castles_2011}, inform social policy \cite{Ratha_2011}, and is a popular topic for public debate \cite{Dempster_Hargrave_2017}. While migration often follows long-term trends driven by development \cite{Skeldon_2008}, it can be dramatically altered by short-term shocks---armed conflict, famine, natural disasters, political instability, changes in national borders, peace agreements, or independence movements \cite{Abel_2019}.

\begin{figure*}[t!]
\begin{adjustbox}{minipage=\textwidth-20pt}
\begin{minipage}{0.5\textwidth}
    \cs{\textbf{A} Registration-based measurements}
\end{minipage}
\begin{minipage}{0.5\textwidth}
    \cs{\textbf{B} Border statistics and passenger surveys}
\end{minipage}
\hfill\vspace{-2mm}

\begin{minipage}{0.5\textwidth}
    \includegraphics[width=\textwidth]{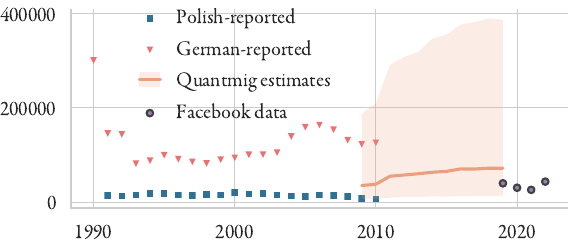}
\end{minipage}
\begin{minipage}{0.5\textwidth}
    \includegraphics[width=\textwidth]{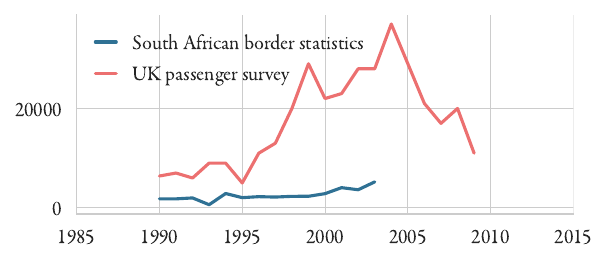}
\end{minipage}
\vspace{0.1em}

\begin{minipage}{0.5\textwidth}
    \cs{\textbf{C} Net migration estimates for France}
\end{minipage}
\begin{minipage}{0.5\textwidth}
    \cs{\textbf{D} Migrant stock: Somalians in Ethiopia}
\end{minipage}
\hfill\vspace{-2mm}

\begin{minipage}{0.5\textwidth}
    \includegraphics[width=\textwidth]{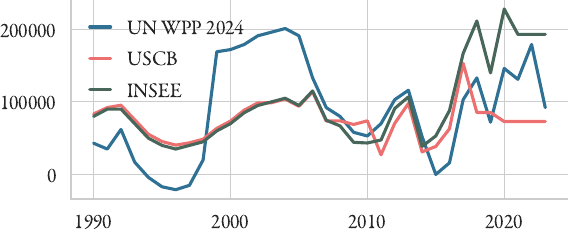}
\end{minipage}
\begin{minipage}{0.5\textwidth}
    \includegraphics[width=\textwidth]{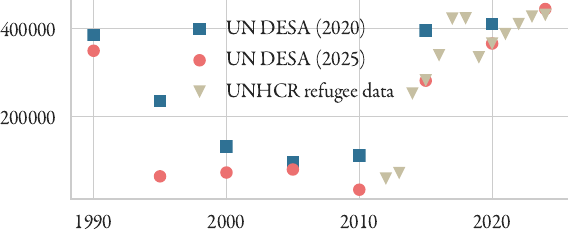}
\end{minipage}
\caption[The difficulty of measuring migration]{The difficulty of measuring migration. \textbf{A} For a small number of countries, migration flow estimates based on domicile registration (usually with a local authority) is available, but the discrepancies can be significant: shown are estimates of flows based on registrations of people arriving from Poland as reported by German authorities (red) and de-registrations of people leaving Poland for Germany, as reported by Polish authorities (blue) \cite{Demig2015}. Also shown are the harmonised Quantmig estimates (orange) \cite{Aristotelous_2022,Smith_Aristotelous_2023} and the recent digital trace estimates based on Facebook data \cite{Chi_Abel_2025}. \textbf{B} Alternative measures such as border statistics and passenger survey estimates are shown here for the South Africa to UK corridor as reported by the origin (blue) and destination (red) \cite{Demig2015}; but the discrepancy is again large. \textbf{C} Various estimates of the net migration for France, such as those from the UN Population Division's World Population Prospects (2024 Revision), the US Census Bureau's International Dataset \cite{USCB_ID}, and INSEE, the French national statistical office \cite{INSEE_data}. \textbf{D} Migrant stock estimates from UN DESA \cite{UN_DESA_Stock} do not agree across revisions. For many countries, they are based on refugee data figures from the UNHCR \cite{UNHCR}.}
\label{fig:net_migration_example_comparison}
\end{adjustbox}
\end{figure*}

Human mobility, however, is notoriously difficult to define and track \cite{Willekens_2016}. A fundamental challenge in monitoring migration patterns lies in defining a migration event: the United Nations recommends a 12-month threshold \cite{UN_migration_definition}, where anyone relocating for the majority of a year or more qualifies as a migrant; but this definition is not applied consistently. Even more difficult is the collection of migration data. Some countries, such as Germany, mandate residential registration, requiring migrants to report their country of origin upon arrival. Others, like the United Kingdom, rely on visa records and passenger surveys. A third common approach uses border entry statistics collected by immigration authorities. Each method has limitations: registration systems typically undercount emigration, since few individuals de-register when leaving; passenger surveys and border data are not comprehensive and may conflate short-term and long-term travellers. As a result, estimates from sending and receiving countries often diverge significantly. In 2005, for instance, Germany reported 160,000 arrivals from Poland, while Poland recorded only 12,300 departures to Germany (see fig. \ref{fig:net_migration_example_comparison}{\cs{\textbf{A}}}–{\cs{\textbf{B}}}). In Europe, to reconcile such discrepancies, statistical demographers have developed models to estimate bilateral migration flows between countries. The most recent study, the QuantMig project \cite{Aristotelous_2022,Smith_Aristotelous_2023}, used a Bayesian framework to estimate bilateral migration flows for 30 European countries from 2009 to 2019. While offering a harmonised approach, the model still reveals substantial uncertainty---in some cases, with credible intervals spanning over 100\%.

Yet these methodological difficulties are overshadowed by a much more fundamental problem: a widespread lack of migration data, particularly across the Global South. In many countries, migration statistics are incomplete or non-existent, and when sudden disasters trigger large-scale movement, reliable data collection becomes impossible. A recent study attempted to bypass these constraints by analysing digital traces \cite{Chi_Abel_2025}. It used monthly Facebook location data to estimate bilateral flows among 180 countries between 2019 and 2022, capturing, for example, the displacement of Ukrainians following the Russian invasion and the Venezuelan migration crisis. Labour migration data represents another important source \cite{ILOStat}, as migrant workers often make up a significant share of international movers. However, here too definitions and data standards vary widely between countries \cite{MigrationDataPortalLabourMigration}, and undocumented migration---by its very nature---remains largely invisible to official statistics.

One macroindicator many countries (especially on the receiving side) are interested in estimating is the \emph{net migration}, the balance of immigration and emigration. A small number of countries publish net migration figures, estimated from immigration and border statistics (see figs. \ref{fig:SI:net_migration} in the SI), while on a global scale, the United Nations Department of Economic and Social Affairs (UN DESA) provides figures from 1950 onwards in its \emph{World Population Prospects} (WPP) reports. These are primarily based on demographic estimates rather than immigration statistics. Since births and deaths are more widely and consistently tracked than migration figures, in principle the net migration can be estimated by subtracting the \emph{natural change} (births minus deaths) from the total population change. While this approach is theoretically sound, in practice it is hindered by irregularities in measuring the total population and its change over time, which are sensitive to inconsistencies in census methodology. Consequently, demographic net migration estimates can differ significantly from immigration-based statistics, even for countries whose population data are of high quality (see fig. \ref{fig:net_migration_example_comparison}{\cs{\textbf{C}}}).

In this work, we combine deep learning with a mechanistic flow model to estimate annual bilateral migration flows for the period 1990--2023 between all 230 countries and regions recognised by the UN. Our data are disaggregated by country of birth, meaning that, besides the flows and the net migration for each country, we also obtain a complete dataset of annual \emph{migrant stocks}, that is, the number of migrants $S_{ij}(t)$ born in country $i$ residing in country $j$ in year $t$. A deep neural network is trained on an extensive set of socio-economic and cultural covariates for each country, allowing us to disentangle the drivers of migration and opening the door to future forecasting of migration flows. The network is trained to match a set of \emph{target data}, comprising the UN DESA migrant stocks, Facebook data, as well as a small number of mainly European bilateral flows and net migration data. The target data are used to construct a \emph{loss function}, which is iteratively minimised during training. Once trained, the neural network acts as a function mapping input covariates to migration flows. By training a family of neural networks and additionally `pushing' the uncertainty on the input data through the network, our approach also enables uncertainty quantification, allowing us to pinpoint the countries where data is inconsistent and collection should be improved. 

This marks a paradigm shift for the computational toolset hitherto used to model migration. Most previous techniques for estimating global migration flows have relied solely on migrant stock data published by the UN DESA, which provides estimates at five-year intervals from 1990 (fig. \ref{fig:net_migration_example_comparison}{\cs{\textbf{D}}}). The simplest estimation techniques are based on \emph{stock differencing} \cite{Abel_Cohen_2019} and assume that the bilateral flow $F_{ij}$ is equal to the difference in stocks $S_{ij}(t+1)-S_{ij}(t)$. Negative differences are either dropped (meaning zero flow) \cite{Beine_2011,Bertoli_Huertas_Moraga_2015} or counted towards flows in the opposite direction \cite{Beine_Parsons_2015}. The simplifying assumption here is that bilateral migration flows only take place from a person's country of birth to a destination; that is, the stock of Swedes in Britain changes \emph{only} due to Swedish people arriving from and returning to Sweden; but not due to Swedish people arriving from, say, Norway. To account for this, a more sophisticated array of so-called \emph{demographic accounting methods} was proposed in \cite{Abel_2013,Abel_Sander_2014,Azose_Raftery_2018}. These attempt to infer a three-dimensional flow matrix $T_{ijk}$, with each entry modelling the flow of people born in $i$ moving from $j$ to $k$, allowing for greater flexibility, but also significantly increasing the number of parameters to be estimated. The flow table is constrained such that its estimates reproduce the stock differences. The stock differences are typically first adjusted to account for births and deaths, such that the estimated flow only reproduces the change in stocks not caused by demographic change.

Stock-based flow estimation approaches all take the stock data at face value; they are also unable to increase the temporal resolution of the estimates, and have thus far only yielded five-year or ten-year flows (the resolution of the UN DESA or World Bank data). These long intervals for estimating migration flows hinder researchers' ability to accurately explain and predict migration flows, particularly when migration dynamics can alter suddenly and briefly. An alternative approach to estimating migration flows is the use of gravity models \cite{Poot_2016}, which are broadly defined as any regression-based approach that relates the flow to a set of \emph{covariates} $\bm{\chi}$. These models can, in principle, capture flows at any resolution, provided the covariates are of sufficient quality and are suitably chosen; however, as shown in \cite{Beyer_2022}, they tend to perform poorly when modelling migration, even with a large and sophisticated set of covariates. The fundamental problem when modelling migration as
\begin{equation*}
    \log T_{ijk}(t) = f(\bm{\chi}_{ijk}(t))
\end{equation*}
is that it represents humans as Markovian, acting \emph{only} based on the current state of the world with no regard to the more distant past. This may be warranted when considering the response to a sudden, cataclysmic event, but is hardly reasonable when incorporating long-term, macro-level political, economic, or social indicators. The decision to leave is, in most cases, not merely predicated on the current economic climate: crises from past years can influence a person's decision, due to a multitude of delayed effects and complex feedback loops. Any model that does not account for the system's \emph{memory} will thus fail to accurately reproduce, let alone explain, the temporal and spatial variance in human migration. In this work, we employ a \emph{recurrent neural network} \cite{Cho_2014,Sutskever_2014}, which implements a form of `memory' by maintaining a `hidden' or `latent' state $\bm{z}(t)$ that changes over time. $\bm{z}$ encodes the system's entire history, and allows modelling the flows of today as a function of, e.g., economic crises from years past.

The article is structured as follows: we first present the estimation results, showcasing the data on a selection of case studies. We validate our method's performance on test data of unseen flows, and compare it to a selection of standard methods discussed above. The inference method is presented in detail in the Methodology section. In the following, we denote by $\bm{S}$ the stock estimates, by $\bm{T}$ the flows disaggregated by birth, by $\bm{F}$ the total origin-destination flows, and by $\bm{\mu}$ the net migration. For notational clarity, we will omit the time argument wherever possible. Estimated quantities will be denoted by a hat, e.g. $\hat{\bm{\mu}}$.

\begin{figure*}[ht!]
\begin{adjustbox}{minipage=\textwidth-20pt}
\begin{minipage}{\textwidth}
    {\cs{\textbf{A}} Largest single-year flow events}
\end{minipage}
\hfill\vspace{-2mm}

\begin{minipage}{\textwidth}
    \includegraphics[width=\textwidth]{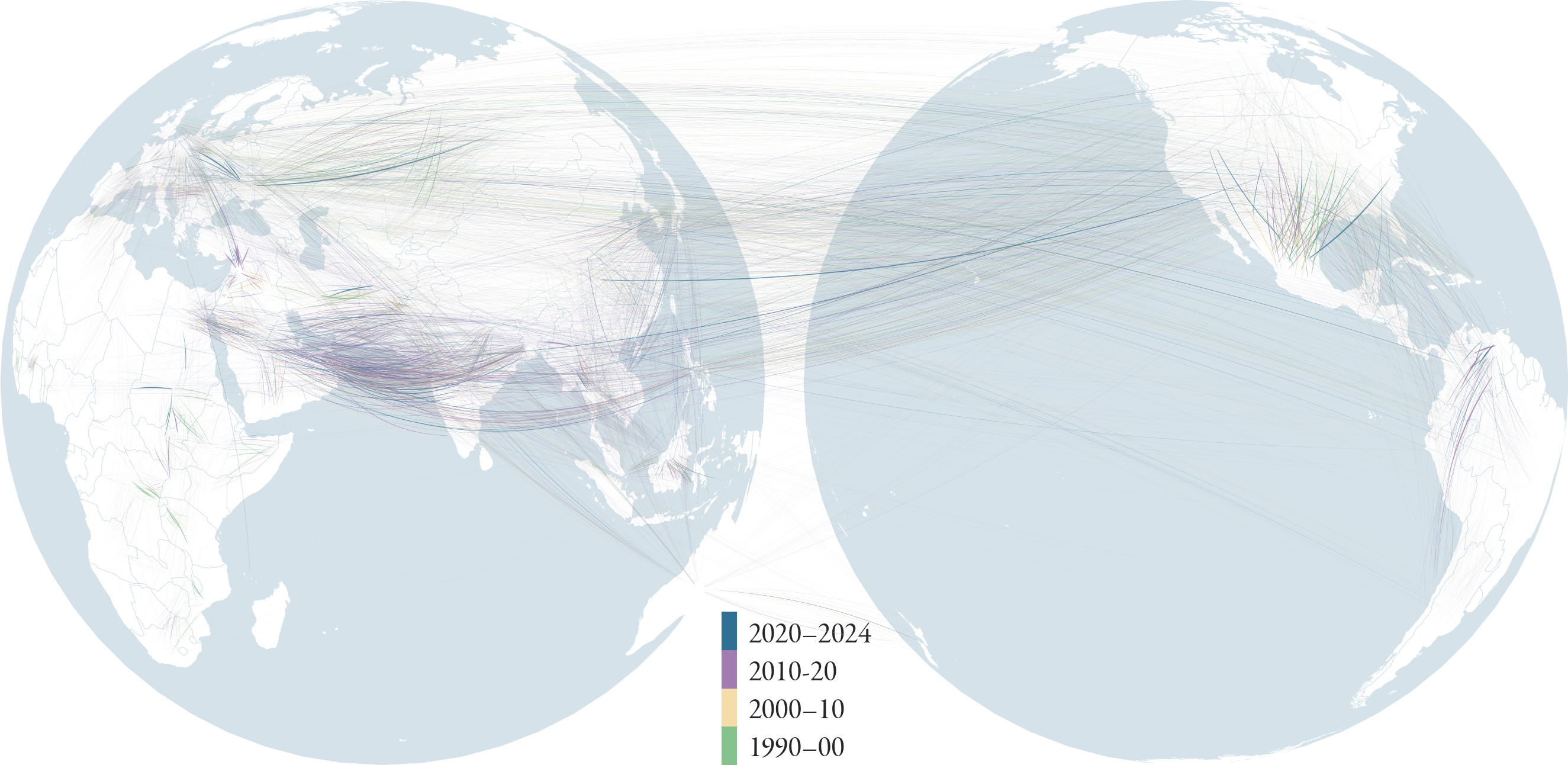}
\end{minipage}
\hfill\vspace{2mm}

\begin{minipage}{\textwidth}
    {\cs{\textbf{B}} Large-scale migration events}
\end{minipage}
\begin{minipage}{\textwidth}
    \includegraphics[width=\textwidth]{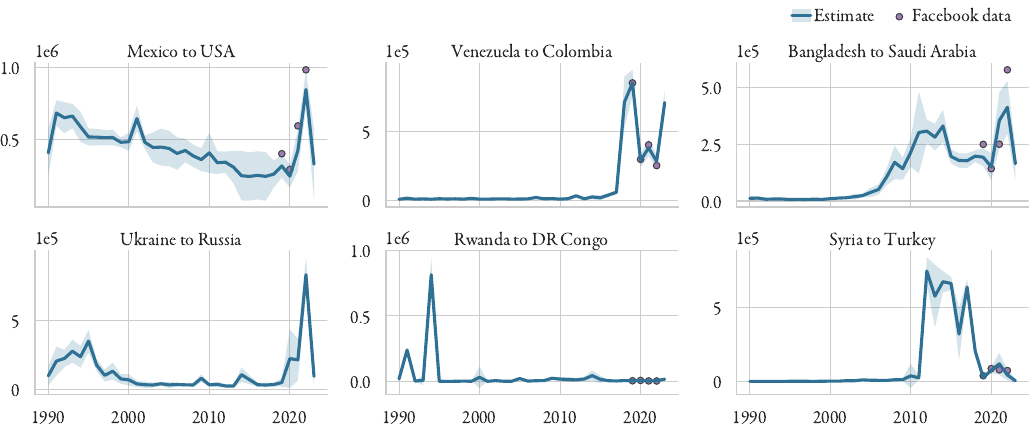}
\end{minipage}
\caption{Our dataset reveals global bilateral, annual migration flows, disaggregated by country of birth, for all countries and territories from 1990--2023. \textbf{A} Shown are the 10,000 largest country-to-country flows $\bm{F}_{ij}$, colour-coded by their decade. Events such as the Rwandan genocide, mass emigration from Venezuela, the Syrian civil war, and Ukrainian emigration are clearly visible. \textbf{B} A selection of large-scale migration events of the past forty years. Also shown is the Facebook data.}
\label{fig:global_flows}
\end{adjustbox}
\end{figure*}
\begin{figure*}[ht!]
\begin{adjustbox}
{minipage=\textwidth-20pt}
\begin{minipage}{\textwidth}
    \includegraphics[width=\textwidth]{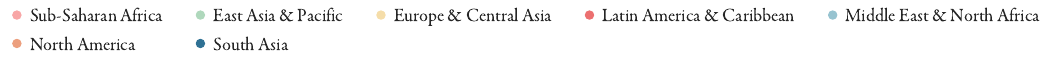}
\end{minipage}

\begin{minipage}{0.33\textwidth}
    \includegraphics[width=\textwidth]{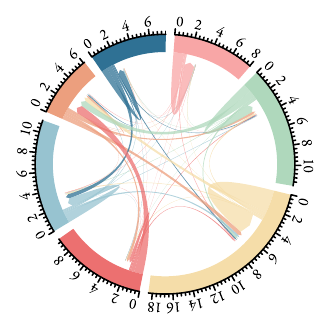}
\end{minipage}
\begin{minipage}{0.33\textwidth}
    \includegraphics[width=\textwidth]{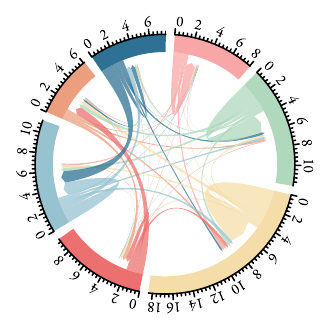}
\end{minipage}
\begin{minipage}{0.33\textwidth}
    \includegraphics[width=\textwidth]{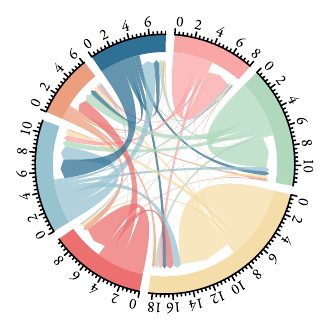}
\end{minipage}
\begin{minipage}{0.33\textwidth}
    \begin{center}{\cs{1990}}\end{center}
\end{minipage}
\begin{minipage}{0.33\textwidth}
    \begin{center}{\cs{2005}}\end{center}
\end{minipage}
\begin{minipage}{0.33\textwidth}
    \begin{center}{\cs{2023}}\end{center}
\end{minipage}
\vspace{0.1em}

\begin{minipage}{\textwidth}
    \includegraphics[width=\textwidth]{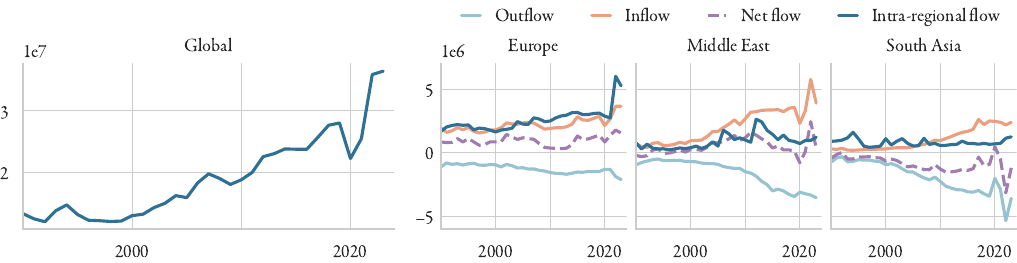}
\end{minipage}
\caption{Regional flow patterns. Top: chord diagrams of global flows for select years. Bottom: total number of international migration movements (left), as well as region-specific in- and outflows and net flows. See figs. \ref{fig:SI:regional_flows}--\ref{fig:SI:regional_flows_per_cap} in the SI for all regions and per-capita values.}
\label{fig:regional_flows}
\end{adjustbox}
\end{figure*}

\begin{figure*}[ht!]
\begin{adjustbox}{minipage=\textwidth-20pt}
\begin{minipage}{\textwidth}
    {\cs{\textbf{A}} Intra-European flows, 1991}
\end{minipage}
\begin{minipage}{\textwidth}
    \includegraphics[width=\textwidth]{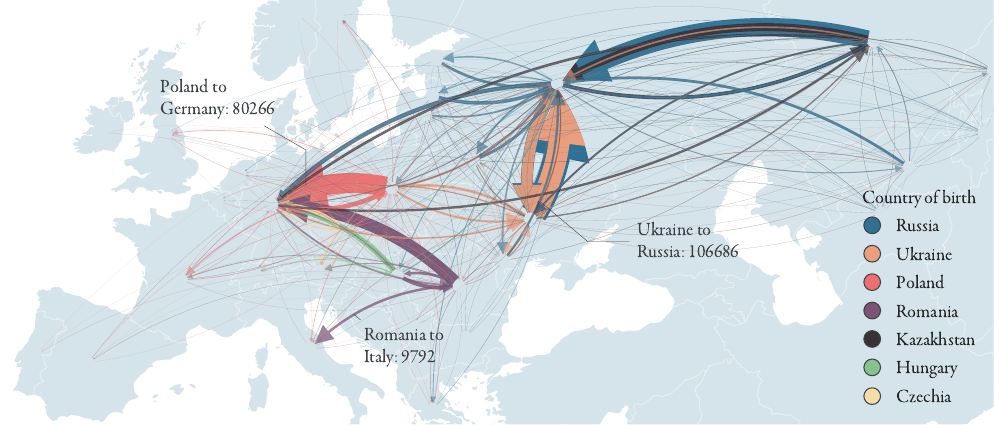}
\end{minipage}
\vspace{0.1em}

\begin{minipage}{\textwidth}
    {\cs{\textbf{B}} Total flows}
\end{minipage}
\begin{minipage}{\textwidth}
    \includegraphics[width=\textwidth]{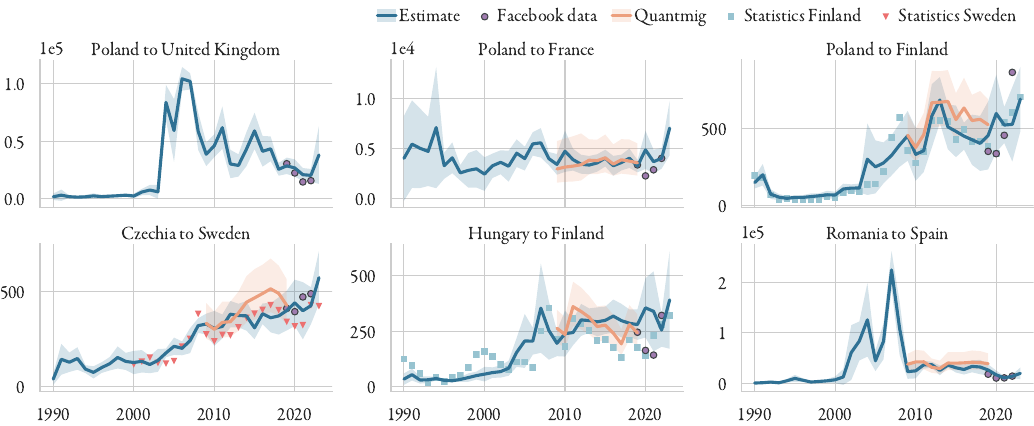}
\end{minipage}

\caption{\textbf{A} Intra-European flows in 1991, colour-coded by country of birth. Some reference flows are indicated for scale. \textbf{B} Total bilateral flows for selected European corridors. Also shown are the estimates from the various target datasets used to train the model (see Methods).}
\label{fig:Europe_1991}
\end{adjustbox}
\end{figure*}

\section{Results}
\paragraph{A global map of migration} Our estimates reveal that, since 2000, global migration movements have risen from 13 million people annually to over 36 million in 2023 (fig. \ref{fig:regional_flows}). This trend is not explained by a rising global population, as per-capita migration saw a similarly steady increase from 0.21\% in 2000 to 0.45\% in 2023 (see figs. \ref{fig:SI:regional_flows_per_cap} in the SI). Since the turn of the millennium, total global migration has only seen two periods of sustained decrease: during the Great Recession in 2008--2009, and during the Covid-19 pandemic in 2020. The largest single-year event we registered is the 2019 movement of people from Venezuela to Colombia, totalling around 850,000, closely followed by migration from Mexico to the US in 2022, and from Ukraine to Russia in 2022. Globally, the Middle East experienced the highest total inflow of migrants, chiefly from South Asia and the Philippines, with immigration from Bangladesh to Saudi Arabia alone averaging around 300,000 people per year from 2010 onwards (see fig. \ref{fig:global_flows}--\ref{fig:regional_flows}). We estimate that, since 2010, a total of 18.4m people, averaging 1.3m per year, migrated from India, Pakistan, and Bangladesh to Saudi Arabia, Qatar, Bahrain, and the UAE---this compares to 14.8m movements from Mexico to the US over the entire period since 1990.  

\begin{figure*}[ht!]
\begin{adjustbox}{minipage=\textwidth-20pt}
\begin{minipage}{\textwidth}
    {\cs{\textbf{A}} Performance on training and test data}
\end{minipage}
\vspace{-2mm}

\begin{minipage}{\textwidth}
    \includegraphics[width=\textwidth]{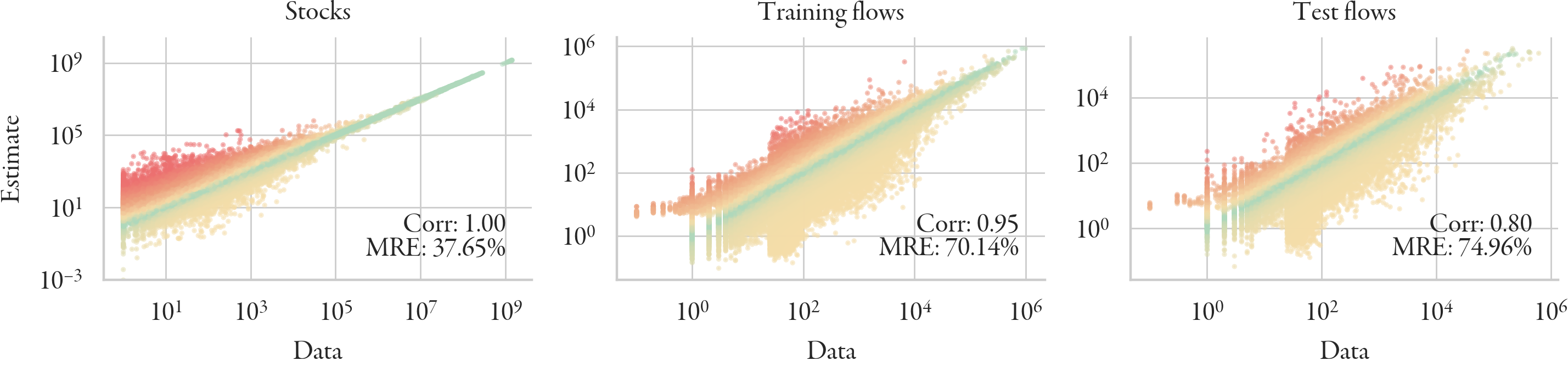}
\end{minipage}

\vspace{2mm}
\begin{minipage}{0.35\textwidth}
    {\cs{\textbf{B}} Correlation on flow corridors}
\end{minipage}
\begin{minipage}{0.65\textwidth}
    {\cs{\textbf{C}} Uncertainty}
\end{minipage}
\vspace{-2mm}

\begin{minipage}{\textwidth}
    \includegraphics[width=\textwidth]{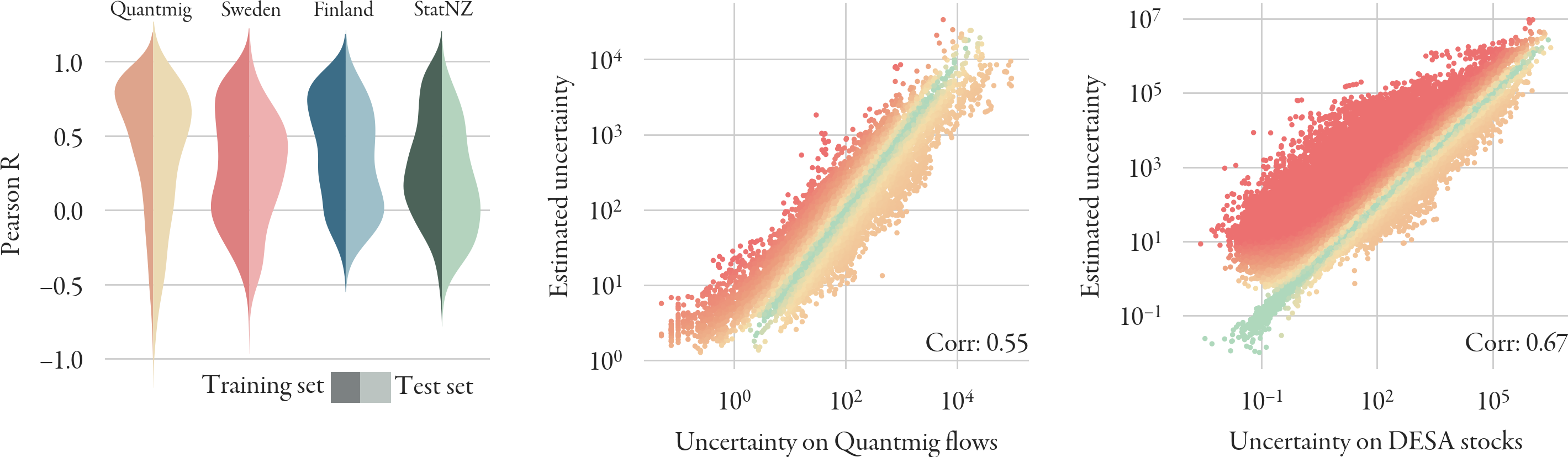}
\end{minipage}

\caption{Performance evaluation of the neural network. \textbf{A} We test the prediction accuracy on a test set of unseen flows, randomly chosen from the target dataset. The network achieves a correlation score of 100\% on the stock data, 95\% on the training flows, and 80\% on the test flows. Also indicated is the median relative error (MRE). \textbf{B} The distribution of correlations along flow corridors, on the training corridors (left half of each violin) and test corridors (right half). Shown are the distributions for the various flow datasets making up the flow target data. \textbf{C} Comparison of the uncertainties on the estimates (y-axes) and the Quantmig flows (left panel) and DESA stocks (right). The uncertainties on the DESA stocks are themselves estimated via demographic accounting and scaling, as described above (see also fig. \ref{fig:SI:stock_weights} in the SI)}
\label{fig:performance}
\end{adjustbox}
\end{figure*}

Europe consistently ranks as the region with the highest volume of intra-regional migration, surpassed only once by Sub-Saharan Africa in the early 1990s during the Rwandan civil war. Pre-2020, gross flows within Europe reached around 3 million people annually, having steadily increased during the 2000s and 2010s following the eastward expansion of the EU and the Schengen region. Flows from Eastern to Western Europe since 1990 total around 20m, or 600,000 per year. Figure \ref{fig:Europe_1991} shows a snapshot of intra-European flows in 1991, following the collapse of the Soviet Union, colour-coded by country of birth. In that year, by our estimates, intra-European flows reached around 2.02m people, of which 807,000 alone were of people born in Poland, Russia, Ukraine, and Romania. The largest movements took place between Ukraine and Russia, Kazakhstan and Russia, and into Germany. During this time, we see significant levels of return migration (bidirectional movement), as some sought to return to their country of birth, while others relocated abroad in search of economic opportunity. Panel {\cs{\textbf{B}}} shows the flow estimates $\hat{\bm{F}}$ for a selection of corridors, alongside values from the various datasets used to train the neural network. Our estimates match not only the data, but also the uncertainty on the Quantmig values exceedingly well (see also fig. \ref{fig:performance}{\cs{\textbf{D}}} and discussion thereon). 

\begin{figure*}[ht!]
\begin{adjustbox}{minipage=\textwidth-20pt}
\begin{minipage}{\textwidth}
    {\cs{\textbf{A}} Flows by country of birth}
\end{minipage}
\hfill\vspace{-2mm}

\begin{minipage}{\textwidth}
    \includegraphics[width=\textwidth]{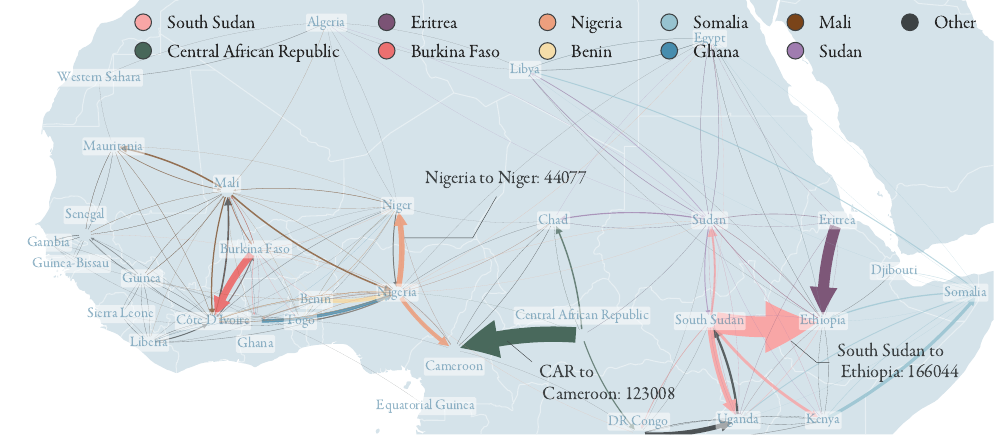}
\end{minipage}

\vspace{3mm}
\hfill\vspace{-2mm}

\begin{minipage}{\textwidth}
    {\cs{\textbf{B}} Migrant stocks}
\end{minipage}
\hfill\vspace{-2mm}

\begin{minipage}{\textwidth}
    \includegraphics[width=\textwidth]{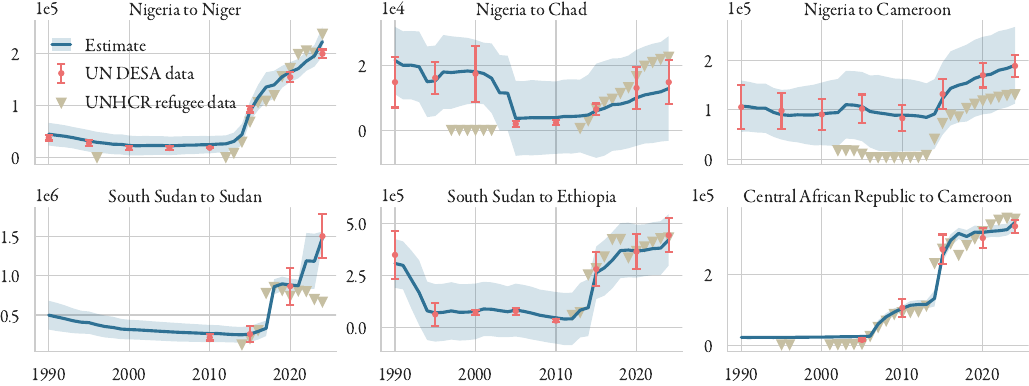}
\end{minipage}

\caption{Migration flows in Sub-Saharan Africa, 2014. \textbf{A} Flows, colour-coded by country of birth. Some reference flows are indicated for scale. \textbf{B} Migrant stocks for selected country pairs. Also shown are refugee figures from the UNHCR.}
\label{fig:Africa_2014}
\end{adjustbox}
\end{figure*}

\begin{figure*}[htp!]
\begin{adjustbox}{minipage=\textwidth-20pt}
\begin{minipage}{\textwidth}
    {\cs{\textbf{A}} Total flows by origin}
\end{minipage}
\hfill\vspace{-2mm}

\begin{minipage}{\textwidth}
    \includegraphics[width=\textwidth]{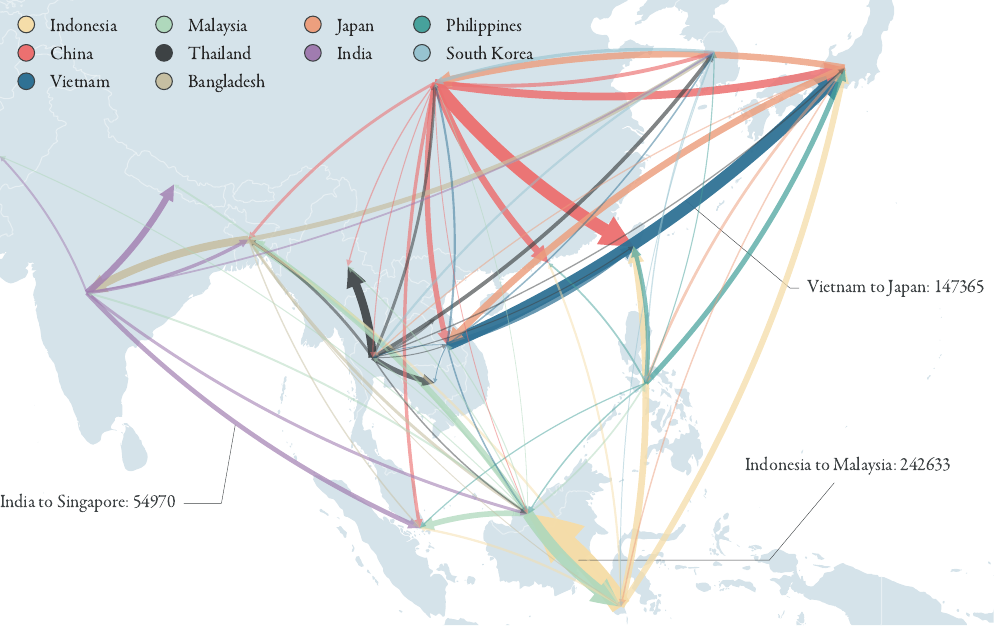}
\end{minipage}

\vspace{3mm}
\hfill\vspace{-2mm}

\begin{minipage}{0.33\textwidth}
    \begin{center}{\cs{\textbf{B}} Migrant stocks}\end{center}
\end{minipage}
\begin{minipage}{0.33\textwidth}
    \begin{center}{\cs{\textbf{C}} Total flow}\end{center}
\end{minipage}
\begin{minipage}{0.33\textwidth}
    \begin{center}{\cs{\textbf{D}} Net migration}\end{center}
\end{minipage}
\hfill\vspace{-2mm}

\begin{minipage}{\textwidth}
    \includegraphics[width=\textwidth]{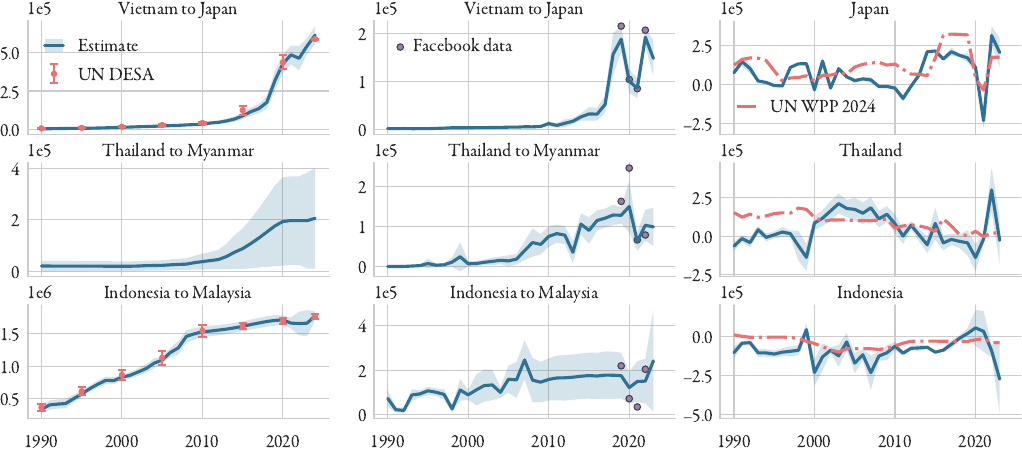}
\end{minipage}
\caption{Migration flows in Asia, 2023. \textbf{A} Shown are the 100 largest origin-destination flows between the given selection of countries, colour-coded by country of origin. \textbf{B}--\textbf{D} Migrant stocks, total origin-destination flows, and net migration estimates for select countries.}
\label{fig:Asia_2023}
\end{adjustbox}
\end{figure*}

\begin{figure*}[ht!]
\begin{adjustbox}{minipage=\textwidth-20pt}
\begin{minipage}{\textwidth}
    {\cs{\textbf{A}} Net migration figures for selected countries}
\end{minipage}
\hfill\vspace{-2mm}

\begin{minipage}{\textwidth}
    \includegraphics[width=\textwidth]{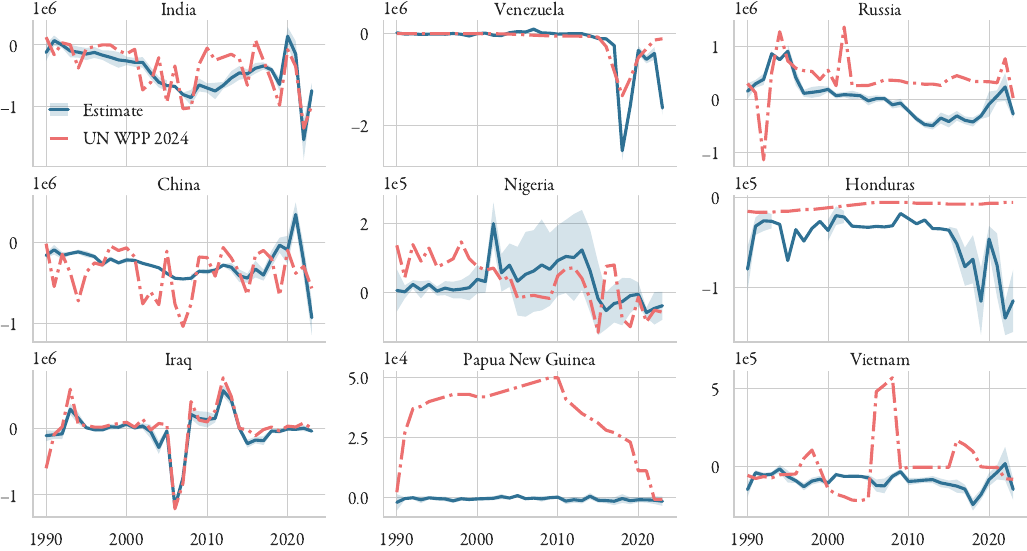}
\end{minipage}

\vspace{5mm}

\begin{minipage}{0.5\textwidth}
    {\cs{\textbf{B}} Correlation with UN WPP net migration data}
\end{minipage}
\begin{minipage}{\textwidth}
    {\cs{\textbf{C}} Median relative uncertainty on net migration estimates}
\end{minipage}
\hfill\vspace{-2mm}

\begin{minipage}{0.5\textwidth}
    \includegraphics[width=\textwidth]{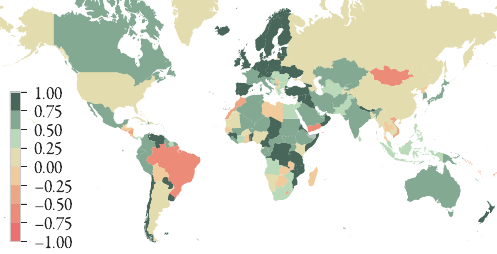}
\end{minipage}
\begin{minipage}{0.5\textwidth}
    \includegraphics[width=\textwidth]{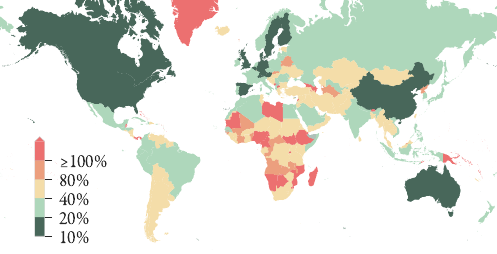}
\end{minipage}
\caption{Net migration estimates and comparison with UN WPP data. \textbf{A} Net migration figures for selected countries, alongside the WPP estimate. \textbf{B} Correlation coefficient of our estimates with WPP figures. \textbf{C} Median relative uncertainty (standard deviation over mean estimate) on our estimates.}
\label{fig:Net_migration}
\end{adjustbox}
\end{figure*}
\paragraph{Migration in the Global South}
Europe is perhaps the region with the least need for a detailed analysis of migratory patterns, since data are (relatively) plentiful. The value of our dataset lies primarily in what it tells us about movements in other parts of the world, especially the Global South. In the mid-2010s, for instance, Sub-Saharan Africa saw several large-scale migration events. Civil war raged in the newly independent country of South Sudan from 2013 onwards, causing a large exodus into neighbouring Ethiopia (fig. \ref{fig:Africa_2014}). The UNHCR classifies the entire migrant population of South Sudanese in Ethiopia as refugees. Violence also erupted in West Africa, with the jihadist group Boko Haram starting an armed insurgency against the Nigerian government in 2009, and dramatically escalating its attacks in 2014, including by abducting nearly 300 young women from a school \cite{CFR_Nigeria_tracker,Guardian_2015}. In 2014 alone, we estimate that around 77,000 persons born in Nigeria moved or fled to neighbouring Chad, Niger, Cameroon, the majority of which (50,000) to Niger. From 2009--2019, we estimate an outflow of Nigerian-born persons to these three countries of 270,000 with a standard deviation of 40,000. This figure is dwarfed by the IOM estimate of around 2.4m internally displaced people as a consequence of the violence \cite{IOM_Nigeria}. Meanwhile, the ongoing civil war in the Central African Republic lead to a continuous outflow to neighbouring Cameroon, DR Congo, and Chad. In East and South-East Asia, home to nearly three-tenths of the worlds population, the largest flows originate from Indonesia and Vietnam, moving to Malaysia, Japan, and Taiwan. The Malaysia-Indonesia corridor is the largest in the region, transporting an estimated 6.6m people from 2000--2024, followed by Thailand-Myanmar (5m in the same period). 

\paragraph{Revising the UN figures} In figure \ref{fig:Asia_2023} we also show the net migration figures for selected countries alongside the estimates from the 2024 WPP report. Our dataset provides a valuable correction to these figures, which, as mentioned in the introduction, are calculated from demographic residuals rather than immigration statistics; the formula used is:
\begin{equation*}
    \mu^\text{WPP}(t) = \Delta P(t) - (\beta(t) - \gamma(t))P(t),
\end{equation*}
with $P(t)$ the total population, and $\beta$ and $\gamma$ the crude birth and death rates, respectively. A further selection of net migration figures is shown in \ref{fig:Net_migration}. The variation in the UNPD figures is often caused by anomalies in the population figures, which strongly affect the change in population $\Delta P$ and cause, for instance, Vietnam's net migration to spike around 2008, only to then fall back to zero in 2010. While the UN figures would suggest positive migrant inflow to Russia since 1995, our estimates show that, in fact, Russian net migration turned negative around 2005, a trend only reversed by the inflow of Ukrainian refugees in 2022. 

\paragraph{Meaningful uncertainty quantification} In figure \ref{fig:Net_migration}{\cs{\textbf{B}}}, we show the correlation between our net migration figures and the most recent UNPD estimates \cite{UN_WPP_2024}. We see a strong positive correlation across the OECD (this is unsurprising as these  countries comprise the target data), as well as across much of the African continent. Our estimates of Indian net migration broadly follow the UNPD trend, but are less erratic; the exodus of workers to the Gulf states, commencing in 2003, is  visible. The net migration estimates for Nigeria, meanwhile, are among the most uncertain of our model predictions: in figure \ref{fig:Net_migration}{\cs{\textbf{D}}} we show the median relative error (standard deviation over mean estimate) for all countries, noting that for Africa, especially sub-Saharan Africa, the uncertainty on the net migration is among the highest in the world. This finding underlines the need for more data collection in that region.

\begin{figure*}[ht!]
\begin{adjustbox}{minipage=\textwidth-20pt} 
\vspace{2mm}
\begin{minipage}{\textwidth}
    {\cs{\textbf{A}} Origin-destination flows}
\end{minipage}
\begin{minipage}{\textwidth}
    \includegraphics[width=\textwidth]{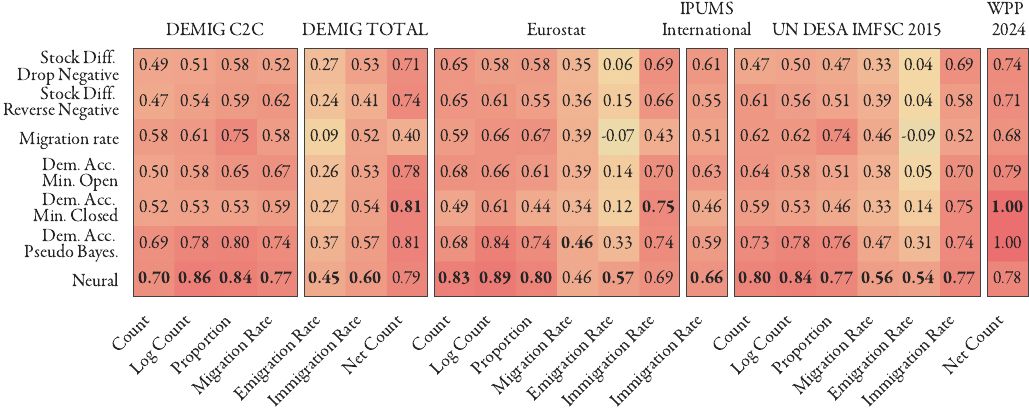}
\end{minipage}
\vspace{2mm}
\begin{minipage}{\textwidth}
    {\cs{\textbf{B}} Birth-destination flows}
\end{minipage}
\vspace{-2mm}

\begin{minipage}{\textwidth}
    \includegraphics[width=\textwidth]{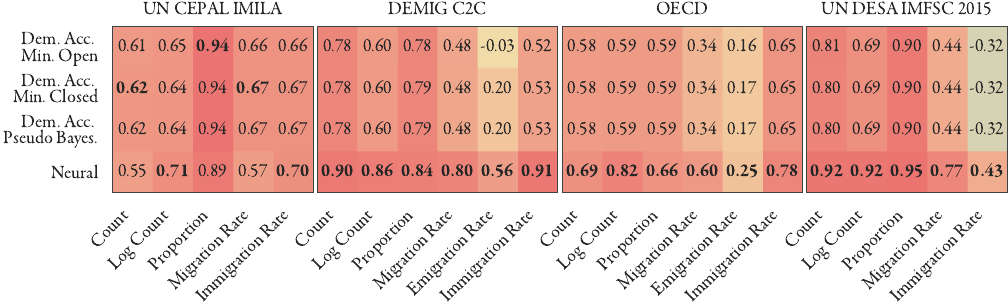}
\end{minipage}
\caption{Comparison of various bilateral flow estimation techniques on unseen collections of migration flow statistics. All values shown are Pearson R coefficients. Since all methods except our own only produce five-year flows; therefore, we aggregate the neural estimates up to five-year intervals for comparison purposes. The evaluation metrics are as in \cite{Abel_Cohen_2019}. \textbf{A} Correlation on origin-destination flows. \textbf{B} Correlation on birth-destination flows.}
\label{fig:comparison}
\end{adjustbox}
\end{figure*}

\section{Testing and Validation}
We validate our approach by testing how well the neural network can reproduce unseen data (the \emph{test data}). To this end, we withhold 20\% of flow corridors from the target dataset and evaluate the trained network on the test flows. Figure \ref{fig:performance}{\cs{\textbf{A}}} shows that it achieves about 95\% correlation on the training data, and 80\% correlation on the test flows, with only a 5\% increase in median relative error (recall that many flows come with considerable uncertainty, and can be small in magnitude, so such a high relative error is not surprising: after all, an estimate of 10 for a flow value of 5 represents a 100\% relative error). While this is the correlation across the entire dataset, we can also examine the distribution of correlations along each corridor (fig. \ref{fig:performance}{\cs{\textbf{B}}}), finding that the neural network generally matches the correlation distributions of the training data on the test set. In figure \ref{fig:performance}{\cs{\textbf{C}}}, we compare the estimated uncertainty of our model with that on the Quantmig data, as well as our estimates of the stock uncertainty. The predicted uncertainty on the flows matches the Quantmig values very well, while producing significantly higher levels of uncertainty on the stocks than obtained through the demographic accounting procedure outlined above.

We further validate the neural predictions on an additional dataset of unseen bilateral flows and compare their performance with those of the various stock-based approaches outlined in the introduction. The datasets and evaluation metrics are presented in \cite{Abel_Cohen_2019,Abel_Cohen_2022} and are broadly comprised of bilateral origin-destination or birthplace-destination flows for a small number of (mostly Western) countries. Since these datasets do not employ a consistent definition of migration, we follow \cite{Abel_Cohen_2019} in not calculating mean errors directly, but rather examining the various correlation metrics across various unilateral and bilateral migration measures commonly used when studying migration patterns and trends. The neural network estimates significantly outperform all other stock-based methods; the only exception are the UN WPP net migration estimates, where the demographic accounting methods, by design, show a perfect correlation of $1$; however, given the methodological issues related to the UN WPP net migration estimates, this is not necessarily desirable.  

\begin{figure}[ht!]
\begin{adjustbox}{minipage=0.49\textwidth-17pt}
    \includegraphics[width=\textwidth]{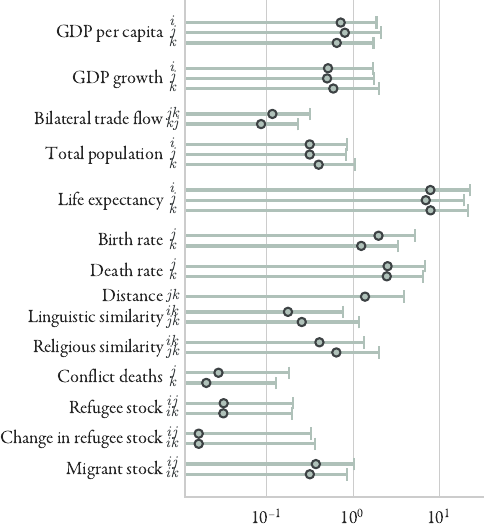}
    \caption{Model elasticities $\nu$ for each continuous covariate. The marker indicates the mean, with the lines indicating one standard deviation (note the log-scaled x-axis). Higher elasticities indicate that the model is more sensitive to changes in that covariate. See also tab.~\ref{table:covariates}}
    \label{fig:elasticity}
\end{adjustbox}
\end{figure}

Lastly, we are interested in gauging how sensitive the model is to the various input covariates. To this end, we calculate the neural network's \emph{elasticity} $\nu$ in each covariate dimension, that is, the relative change induced in $T_{ijk} \equiv T$ by a relative change in the $i$-th covariate:
\begin{equation}
    \nu_i = \left\vert \dfrac{\chi_i}{{T}} \dfrac{\partial {T}}{\partial {\chi_i}}\right\vert = \left\vert \dfrac{\partial \log {T}}{\partial{\chi_i}}\right\vert \vert \chi_i \vert.
    \label{eq:elasticity}
\end{equation}
Figure \ref{fig:elasticity} shows that, on average, the model predictions are most sensitive to changes in the life expectancy and mortality rates, presumably since these act as proxies for a country's (relative) quality of life. Of the economic covariates, GDP per capita is the most determinant; religious similarity appears more determinant than linguistic similarity, and data on conflict and refugee stocks is the least determinant.

\section{Discussion}
Data on human movements is notoriously unreliable, noisy, and---crucially---absent for large parts of the world. Our work significantly increases the spatial and temporal resolution of existing migration data. We provide a complete set of annual migrant stock values, net migration estimates for each country, and bilateral flow estimates disaggregated by birth country. Reliable migration data are of great interest to economists for use as covariates in econometric models, to epidemiologists studying the spread of infectious diseases, and to demographers creating population growth projections. Our dataset---along with the supplementary training data, including real GDP estimates for 230 countries and territories---will be a valuable asset. More importantly, this study provides a foundational tool for breaching the next frontiers in migration modelling. For one, researchers are increasingly seeking to move beyond country-level migration flows toward more geographically granular modelling. Many of the largest population movements occur within countries, rather than between them \cite{Rees_2017}. Replacing countries with a finer spatial grid as the fundamental unit would represent a significant advance in statistical demography. Our framework is well-suited to support this transition. A recent high-resolution dataset of demographic data at a 10km\textsuperscript{2} resolution was released \cite{Venla_2023}, which could be used to train the network. A similar recurrent architecture would be required to capture spatial correlations, allowing the outflow from one cell to influence movements in distant neighbouring regions. As shown, deep neural networks are generally adept at learning such long-range, complex correlations in both space and time---including through more advanced transformer architectures and graph neural networks \cite{Bronstein_2017, GNNBook2022}. These architectures may further expand the capabilities of our method in scenarios with intricate spatial structure. Additionally, as we have demonstrated, machine learning methods can bring together multiple sources of migration data and measures. Combining data enables more robust estimates of migration, leveraging the benefits of both traditional data sources on migration and newer sources based on digital trace data. Finally, deep learning methods capture complex non-linear trends and threshold effects, which can be utilised when predicting or creating scenario-based projections of future migration patterns. Without a strong theoretical basis, shifts in future migration patterns are often overlooked in population forecasting exercises, relying instead on assumptions that follow long-term trends or converge to zero to mitigate the impact of migration on future population sizes \cite{Abel_2018}. However, migration is becoming the most significant component of population change, which will be better served with modelling frameworks that can incorporate future human system responses to as yet unforeseen events, such as climate changes or new conflicts in previously peaceful areas. 

\vspace{1em}
\hrule

\paragraph{Data availability} The complete flow datasets are available at \href{https://doi.org/10.5281/zenodo.15623215}{doi.org/10.5281/zenodo.15623215}. A trained neural network, all training covariates, the code used to train the network, as well as a Jupyter notebook to evaluate the models and reproduce the plots in this publication, are available at \href{https://github.com/ThGaskin/Migration_flows}{github.com/ThGaskin/Migration{\textunderscore}flows}. Detailed descriptions on data collection are provided in the README files, as well as the Supplementary Material.

\paragraph{Acknowledgements} The authors would like to thank Robert Beyer (International Organization for Migration) for useful discussions on the inference procedure, and for suggesting the use of demographic accounting to estimate the uncertainty on the stock target data.

\vspace{1em}
\hrule
\newpage

\begin{figure*}[ht!]
\begin{adjustbox}{minipage=\textwidth-20pt}
\begin{minipage}{\textwidth}
    \cs{\textbf{A} Training method}
\end{minipage}
\begin{minipage}{\textwidth}
\begin{picture}(\textwidth, 0.42\textwidth)
    \put(0, 0){\includegraphics[width=0.95\textwidth]{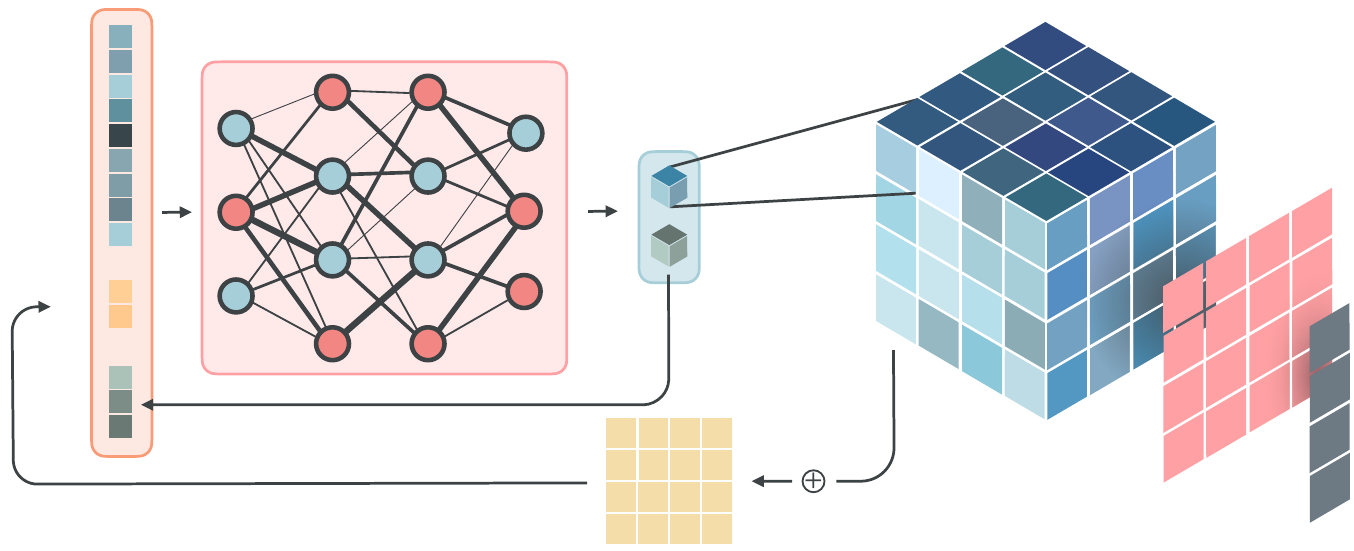}}

    \put(32, 187){\cs{\it Input}}
    \put(20,130){\rotatebox{90}{\cs{$\bm{\chi}_{ijk}(t)$}}}
    \put(20,70){\rotatebox{90}{\cs{ $\psi_\lambda(\bm{S})$}}}
    \put(20,37){\rotatebox{90}{\cs{ $\bm{z}_{ijk}(t)$}}}

    \put(100,168){\cs{\it Neural network $u_\theta$}}

    \put(212,140){\cs{\it Output}}
    \put(202,125){\cs{$\hat{T}_{ijk}$}}
    \put(233,98){\cs{\it Latent state}}
    \put(233,90){\cs{$\bm{z}_{ijk}(t+1)$}}

    \put(400,160){\cs{{\it Flow table} $\bm{T}(t)$}}
    \put(320,35){\cs{\it Origin-destination}}
    \put(320,25){\cs{{\it flow} $\bm{F} = \sum_i T_{ijk}$}}
    \put(390,12){\cs{\it Net migration}}
    \put(390,1){\cs{$\bm{\mu} = \sum_i F_{ij} - F_{ji}$}}
    \put(146,10){\cs{{\it Stocks} $\bm{S}(t+1)$}}
\end{picture}
\end{minipage}
\vspace{0.3em}

\begin{minipage}{0.53\textwidth}
    \cs{\textbf{B} Yeo-Johnson data transformation $\psi_\lambda$}
\end{minipage}
\begin{minipage}{0.47\textwidth}
    \cs{\textbf{C} Constructing the training loss}
\end{minipage}
\hfill\vspace{-2mm}

\begin{minipage}{\textwidth}
    \begin{picture}(\textwidth, 0.25\textwidth)
        \put(0, ){\includegraphics[width=\textwidth]{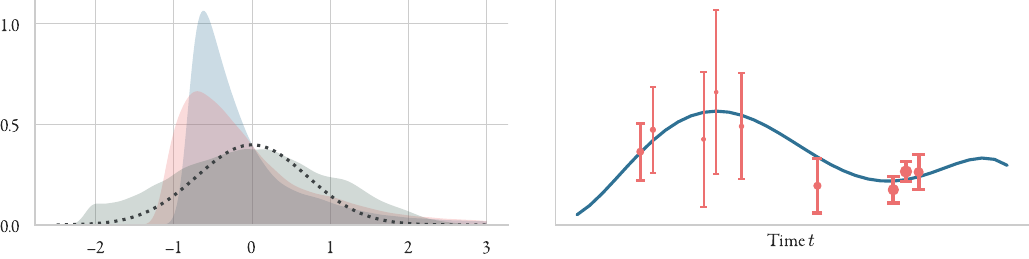}}
        \put(72, 85){\cs{\textcolor{c_darkblue}{$\psi_{\lambda_1}$}}}
         \put(62, 53){\cs{\textcolor{c_red}{$\psi_{\lambda_2}$}}}
         \put(50, 33){\cs{\textcolor{c_darkgreen}{$\psi_{\lambda_3}$}}}
         \put(133, 52){\cs{\textcolor{c_darkgrey}{$\mathcal{N}_0$}}}
        \put(355, 100){\cs{$J \sim \sum w_k (\psi_\lambda(\hat{y}_k) -\psi_\lambda(y_k))^2$}}
        \put(300, 100){\cs{\textcolor{c_red}{$y\pm\Delta y$}}}
        \put(385, 50){\cs{\textcolor{c_darkblue}{$\hat{y}$}}}
    \end{picture}
\end{minipage}
\vspace{-2mm}
\caption{Overview of the training method. \textbf{A} For each edge, the covariates $\bm{\chi}_{ijk}$, the migrant stock $(S_{ij}, S_{ik})$, and the latent state $\bm{z}_{ijk}$ are passed through the neural network $u_\theta$ to produce an estimated flow $\hat{T}_{ijk}$ and an updated latent variable $\bm{z}_{ijk}(t+1)$. The complete flow table $\bm{T}$ is assembled from all the individual estimates and used to calculate the net migration $\bm{\mu}$, the origin-destination flow $\bm{F}$, and the migrant stock $\bm{S}(t+1)$ of the next year, which is then fed back into the neural network as input. This gives a recursive training procedure. \textbf{B}~Input and target data are transformed to more closely resemble a normal distribution (dashed line) using a Yeo-Johnson type transformation (eq.~\eqref{eq:yeo_johnson}). This is done by tuning the parameter $\lambda$ using maximum likelihood estimation. \textbf{C} The training loss $J$ is calculated from the residuals of the prediction and the observation data. Each term is weighted by the uncertainty on the target value, with smaller uncertainties leading to a higher weight. }
\label{fig:method}
\end{adjustbox}
\end{figure*}

\newpage
\section{Methodology}
UN DESA provides estimates of global migrant stock $S_{ij}(t)$, that is, the number of people born in country $i$ living in country $j$ at time $t$ \cite{UN_DESA_Stock}. This data are given at five-year intervals from 1990 to 2020, as well as a recent estimate for 2024. The stocks evolve according to the equation 
\begin{align}
    \partial_tS_{ij}(t) = & \underbrace{\delta_{ij}B_j(t)}_{\text{births}} - \underbrace{\gamma_j(t)S_{ij}(t)}_{\text{deaths}} \nonumber \\ & + \underbrace{\sum_k (T_{ikj} - T_{ijk})(t)}_{\text{migration}},
    \label{eq:stock_evolution}
\end{align}
where $B_j$ and $\gamma_j$ are, respectively, the total number of births and the mortality rate of the destination, and $\delta_{ij}$ is the Kronecker matrix 
\begin{equation}
    \delta_{ij} = \begin{cases}1, \ \text{if} \ i =j, \\ 0, \ \text{else.}\end{cases} \label{eq:Kronecker_delta}
\end{equation} 
The first term in eq. \eqref{eq:stock_evolution} simply means that all births in a country increase the native-born stock $S_{ii}$; the second models population decrease due to deaths; the third models the change in stocks due to migration. $T_{ijk}$ is the \emph{flow} of migrants born in $i$ from country $j$ to $k$, and is the quantity we wish to infer. The total flow of people from $j$ to $k$, regardless of their place of birth, is of course then given by
\begin{equation}
    F_{jk} = \sum_i T_{ijk},
    \label{eq:flow_by_origin}
\end{equation}
while a country's \emph{net migration} (arrivals minus departures) is
\begin{equation}
    \mu_j = \sum_k F_{kj} - F_{jk}.
\end{equation}

\subsection{Target data}
Aside from the stock data, there are numerous datasets of partial observations of the flow $\bm{F}$ and the net migration $\bm{\mu}$ to fwhich we could constrain our estimate $\hat{\bm{T}}$, though as mentioned in the introduction, these do not always use consistent definitions of a migration event. The UN DESA World Population Prospects dataset \cite{UN_WPP_2024} and the United States Census Bureau International Database \cite{USCB_ID} both provide estimates of annual net migration for all countries from 1950--2024. These figures are mainly calculated as the residual between the total change in population and natural growth (births minus deaths), and for most countries, they are not derived from immigration statistics; we thus do not include them as target variables. Instead, we use net annual migration statistics from around 30 countries and territories in Europe, North America, Oceania, and East Asia, sourced from national statistical bureaus (see SI).

\begin{figure*}[ht!]
\begin{adjustbox}{minipage=\textwidth-20pt}
\begin{minipage}{0.5\textwidth}
    \cs{\textbf{A} Real GDP per capita, 2019 (const. 2015 USD)}
\end{minipage}
\begin{minipage}{0.5\textwidth}
    \cs{\textbf{B} Religious similarity (Nigeria)}
\end{minipage}
\hfill\vspace{-2mm}

\begin{minipage}{0.5\textwidth}
    \includegraphics[width=\textwidth]{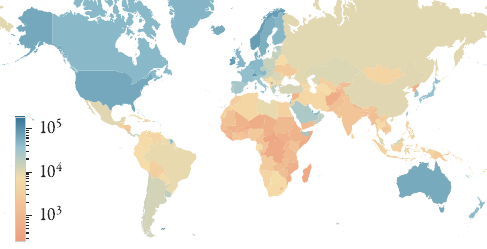}
\end{minipage}
\begin{minipage}{0.5\textwidth}
    \includegraphics[width=\textwidth]{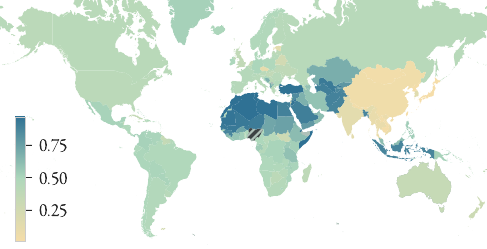}
\end{minipage}
\caption{Two covariates used to train the model. \textbf{A} Real GDP per capita, in constant 2015 USD, as of 2019. \textbf{B} The bilateral religious similarity score for Nigeria, measuring the overlap in adherents of major religions with other countries.}
\label{fig:covariates}
\end{adjustbox}
\end{figure*}

Observations of total origin-destination flows $\bm{F}$ are taken mainly from four sources, which all employ a one-year definition of migration flows:
\begin{itemize}
    \item Harmonised intra-European flows: the Quantmig database \cite{Aristotelous_2022,Smith_Aristotelous_2023} provides probabilistic estimates of migration flows between 30 countries in Europe from 2009--2019, based on publicly-available Eurostat data. These have been harmonised to use a common definition of migration, and also provide uncertainty estimates, which we use to weight the target data points in the loss function used to train the neural network (see below).
    \item National immigration statistics from Sweden, New Zealand, and Finland \cite{Statistics_Sweden_flows,Stat_NZ_flows,Statistics_Finland_flows}: all three countries report total annual in- and out-flows by origin and destination.
    \item Facebook data \cite{Chi_Abel_2025}: estimates of annual bilateral migration flows between 181 countries from 2019--2022 from an analysis of online social media data. We only include annual flows of at least 25 people, since noise was added by the authors to prevent data disclosure which distorts small values.
\end{itemize}
Target values are prioritised in this order, meaning that if two datasets both contain values for the same origin-destination pair, we use the source furthest up in the list. 

\subsection{Input covariates}
Each value $T_{ijk}$ is a flow through a network multi-edge connecting the birth country $i$, the origin $j$, and the destination $k$. We train a deep neural network to learn a mapping $\bm{\chi}_{ijk}(t) \mapsto \hat{T}_{ijk}(t)$, where $\bm{\chi}$ is a vector of economic, social, and geographic covariates pertaining to the three connected countries. In the following we give a summary of the covariates used; see tab. \ref{table:covariates} for an overview, and the Supplementary Information for further details.

\paragraph{Demographic covariates} For each country $i$, $j$, $k$, we include the total population and life expectancies; for the origin and destination countries, we also include birth and death rates, all taken from the UN WPP dataset \cite{UN_WPP_2024}. 

\paragraph{Economic covariates} For each country $i$, $j$, $k$, we include annual real GDP per capita (in constant 2015 US Dollars) and annual GDP growth rate (in \%). data are taken from the World Bank and UNCTAD \cite{WB_GDP_PCAP,WB_GDP_growth,UNCTAD_GDP}, as well as national statistical bureaus. Missing values are calculated by deflating nominal to real GDP using the World Bank deflator; where the deflator is not given, we estimate the deflator from neighbouring or similar nations. Gaps are also filled by calculating the GDP growth rate from the Maddison Project dataset \cite{Maddison_GDP,Bolt_2024}, which provides GDP in constant 2011 purchasing power parity. The growth rates are then used to extrapolate GDP back or forward in time. We also input bilateral trade flows between origin and destination, given in real 2015 USD. These are mostly taken from the harmonised BACI dataset \cite{BACI,BACI_dataset}, and missing values are again extrapolated using the growth rates from the UN Comtrade and IMF Direction of Trade datasets \cite{CEPII}. 

\paragraph{Geographic, cultural, and political covariates}
Religious and linguistic proximity, measuring the extent of overlap in religious affiliation and common spoken languages, are taken from the Correlates of War World Religion Data and USITC Domestic and International Common Language Database datasets \cite{Correlates_of_war_religion, USITC_DICL}. Religious similarity measures the overlap in the number of adherents of major religions; for two countries with religious makeup $\bm{\alpha}_i$ and $\bm{\alpha}_j$, the similarity score is simply the dot product $\bm{\alpha}_i \cdot \bm{\alpha}_j$. The linguistic similarity score is given by the `common spoken language' index; missing entries are filled with the average of linguistic proximity and common native language \cite{USITC_DICL}. We also include the population-weighted geodesic distance \cite{CEPII}, as well as a number of binary covariates: EU membership of the three indexed nations; binary variables $\text{Colony}_{ik}$ and $\text{Colony}_{jk}$, which are $1$ if the first indexed country was ever a colony of the second; as well as the two binary variables $\delta_{ij}$ and $\delta_{ik}$, with $\delta$ the Kronecker delta (eq. \eqref{eq:Kronecker_delta}), which indicate whether an individual is a native of the origin or the destination. 

\paragraph{Conflict deaths, refugee and migrant stock}
To model short-term, disaster-driven migration, we include data on wars and other shocks that trigger large population movements. We include the total number of deaths related to organised violence in the origin and destination countries, given by the Uppsala Conflict Data programme's Georeferenced Event Dataset \cite{UCDP_1,UCDP_2}. We include the total number of refugees, asylum seekers, and others in need of international protection, as given by the UNHCR's refugee statistics \cite{UNHCR}, and also include the annual change in the figures. Lastly, we also input the total migrant stock both in the origin and the destination, $S_{ij}(t)$ and $S_{ik}(t)$, for each year. Since the UN only provides these data points at most every five years, intermediate values are taken from the neural network prediction $\hat{\bm{S}}$ itself. Where initial values are missing, we extrapolate the stocks back to 1990 using a weighted average of similar countries; the weights are calculated by considering correlations across time and space.

\begin{table}[t!]
\flushleft
\fontsize{7pt}{7pt}\selectfont\tablefont
\begin{tabularx}{0.49\textwidth}{ 
   | >{\arraybackslash\hsize=.5\hsize}X| 
   >{\centering\arraybackslash\hsize=.15\hsize}X|
   >{\arraybackslash\hsize=.35\hsize}X|}
  \hline
  \rowcolor{lightblue}
  \textbf{Covariate} & \textbf{Indexed nations} & \textbf{Source} \\
 \hline
Total population & B,O,D & \cite{UN_WPP_2024} \\
\rowcolor{lightblue}
Life expectancy at birth & B,O,D & \cite{UN_WPP_2024}  \\
Birth rate & O,D & \cite{UN_WPP_2024}  \\
\rowcolor{lightblue}
Death rate & O,D & \cite{UN_WPP_2024}  \\
GDP per capita (2015 USD)& B,O,D & \cite{WB_GDP_PCAP,UNCTAD_GDP}, other: see SI \\
\rowcolor{lightblue}
GDP \% annual real growth & B,O,D& \cite{WB_GDP_growth,Maddison_GDP,IMF_WEO_2024}, other: see SI\\
Bilateral trade flow (2015 USD) & OD, DO & \cite{BACI,CEPII}; see SI \\
\rowcolor{lightblue}
Distance & OD & \cite{CEPII}  \\
Religious similarity & BD, OD & \cite{Correlates_of_war_religion,CIA_Worldfactbook}  \\
\rowcolor{lightblue}
Linguistic similarity & BD, OD & \cite{USITC_DICL}  \\
EU membership & B,O,D & \cite{CEPII}  \\
\rowcolor{lightblue}
Colonial relations & BD, OD & \cite{USITC_DGD}  \\
Conflict deaths & O,D & \cite{UCDP_1,UCDP_2}  \\
\rowcolor{lightblue}
Refugee stock & BO, BD & \cite{UNHCR}  \\
Change in refugee stock & BO, BD & \cite{UNHCR}  \\
\rowcolor{lightblue}
Migrant stock & BO, BD & Predicted and \cite{UN_DESA_Stock}  \\
Native of origin $\delta_{ij}$ & BO &   \\
\rowcolor{lightblue}
Native of destination $\delta_{ik}$ & BD &   \\
\hline
\end{tabularx}
\caption{Input covariates $\chi_{ijk}$ used to train the model. The `indexed nations' column indicates for which indices data are included, i.e. place of birth (B), origin (O), or destination (D). Bilateral variables are naturally indexed by two places.}
\label{table:covariates}
\vspace{3mm}\hrule
\end{table}

\subsection{Training}
We apply the training methodology broadly outlined in \cite{Gaskin_2023}. We wish to not only incorporate information about the current state of the world at time $t$, contained in $\bm{\chi}_{ijk}(t)$, but also about the past. To do this, we use a \emph{recurrent} neural network $u_\theta$, which takes as input the covariates (including stocks) as well as a $Z$-dimensional \emph{hidden or latent} state $\bm{z}_{ijk}(t)$. This latent variable represents a `memory' of past changes and their effects on the present flow estimate. The neural network outputs a (log-scaled) estimate of the flow $T_{ijk}$, as well as the updated latent state $\bm{z}_{ijk}(t+1)$, which is then input to the neural network to predict the next point in time:
\begin{equation}
    u_\theta(\bm{\chi}_{ijk}, \bm{z}_{ijk}) = (\log T_{ijk}, \bm{z}_{ijk}(t+1) ) \in \mathbb{R}^{1+Z}.
    \label{eq:nn_output}
\end{equation}
Note that the estimates $\hat{T}_{ijk}$ and all their derived quantities will be real-valued, despite integer target data. This gives a recursive training procedure, where each output is fed back into the neural network to inform the next estimate (see fig.~\ref{fig:method}{\cs{\textbf{A}}}). The latent state $\bm{z}_{ijk}$ is initialised at zero and can take any value in $\mathbb{R}^Z$.

The neural network consists of a set of trainable parameters $\bm{\theta}$ that are optimised using the gradient of the \emph{loss function} $J$. $J$ is designed to ensure that predicted and observed stocks, net migration values, and flows $F_{jk}$ agree, and is structurally an $L^2$-loss of all the different values $\{ y_k \}$. We make two important modifications to this basic loss function: first, we scale the data to make the errors $\hat{y} - y$ more normal, ensuring the loss function is not dominated by the largest values (this will be addressed below); and secondly, we weight each term in the loss function by its \emph{uncertainty} to bias the loss toward values in which we have greater confidence.:
\begin{equation}
    J_\theta \sim \sum_k w_k (\hat{y}_k - y_k)^2,
    \label{eq:weighted_loss}
\end{equation}
with the index $k$ ranging over all the target values in a single \emph{batch}.
The weights $w_k$ are constructed from the relative uncertainty on each point, clamped to the interval $[0.5, 2]$, and normalised such that the mean weight is $1$. The Quantmig dataset provides standard errors on the estimates we use to populate the weights for the flow targets; for all other flow targets, we set the weight to the average weight of the Quantmig weights or 1. For the stocks, we apply the demographic accounting scheme presented in \cite{Abel_Sander_2014}: given the stock tables for two successive years $\bm{S}(t_1)$, $\bm{S}(t_2)$, we add births and deaths and constrain the resulting tables to match their midpoint marginals using iterative proportional fitting. Subsequently subtracting births and deaths again gives two demographically balanced stock estimates for each year, from which we can estimate the uncertainty on each value $S_{ij}$. For the net migration targets, we set the weights to $1$ (see SI for details).

\subsection{Scaling the input and target data}
Much of the input and target data are heavily skewed Poisson distributions, with long tails caused by a small number of strong outliers; in order to improve learning, it is common practice to transform data to make it more normal. To this end we use a symmetrised Yeo-Johnson transform:
\begin{equation}
    \psi_\lambda (x) =  \mathrm{sgn}(x) \times \begin{cases}  \dfrac{(\vert x \vert+1)^\lambda - 1}{\lambda} \ \text{if} \ \lambda \neq 0 \\ \log(\vert x \vert+1) \ \text{else} \end{cases},
    \label{eq:yeo_johnson}
\end{equation}
with $\mathrm{sgn}(x)$ the sign function. Compared to the standard transformation \cite{Yeo-Johnson}, the symmetrisation ensures that negative values are transformed more evenly. The parameter $\lambda$ can be chosen to move the distribution of $\psi_\lambda(x)$ closer towards a normal distribution (see fig. \ref{fig:method}{\cs{\textbf{B}}}); for $\lambda=1$, $\psi$ is simply the identity. The transformed input data are further normalised to have zero mean and unit variance. Note that the transformation eq. \eqref{eq:yeo_johnson} is invertible, with inverse $\psi^{-1}_\lambda$; we can thus always reverse any transformation to obtain the original data. We rescale all non-binary covariates except the religious and linguistic similarity indices to be approximately normal (see SI for the $\lambda$ values used for each). 

To improve prediction accuracy on edges with smaller flows, we also transform the target data using the above function $\psi$; the loss function then reads
\begin{align}
    J_\theta = & \left\langle w^s_{ij}\left( \psi_{\lambda_1}(\Delta \hat{S}_{ij}) - \psi_{\lambda_1}(\Delta S_{ij}) \right)^2 \right\rangle \nonumber \\
    + & \left\langle w^\mu_{ij}\left( \psi_{\lambda_2}(\hat{\mu}_{j}) - \psi_{\lambda_2}(\mu_{j}) \right)^2 \right\rangle \nonumber \\
    + & \left\langle w^f_{ij}\left( \psi_{\lambda_3}(\hat{F}_{ij}) - \psi_{\lambda_3}(F_{ij}) \right)^2 \right\rangle.
\end{align}
Here, $\langle \cdot \rangle$ denotes the average over all target values. Observe that we are not matching stock values directly, but rather the difference in stocks over five-year intervals. This is to avoid conditioning the stock value on (possibly erroneous) initial values, and ensure independence of the stock estimates. An optimal initial stock value can be estimated after training using a least squares approach, in order to fit the time series $\hat{S}_{ij}(t)$ to the data (see below).

\subsection{Model selection and validation}
To select the architecture of the neural network, that is the number of neurons and layers, the activation functions, and the latent dimension $Z$, we use hyperparameter tuning on synthetic data (see SI for details). We use a deep network with 7 layers, 60 neurons per layer, and a hyperbolic tangent on each layer except the last, where we use the CeLU function \cite{torch_CELU}
\begin{equation*}
    \max(0, x) + \min(0, \alpha (\exp(x/\alpha)-1))
\end{equation*}
with $\alpha=-12$. The latent dimension $Z$ was set to $100$. Also by using a hyperparameter sweep on synthetic data, the scaling parameters for the target data $\lambda_i$ were all set to $0.7$.

\subsection{Uncertainty quantification}
Uncertainty on the estimates stems from two sources: first, the degree to which the inference problem is \emph{ill-posed}, meaning the number of possible solutions that fit the data; and second, the uncertainty on the input covariates themselves \cite{Gaskin_2023}. The uncertainty arising from the (potential) ill-posedness of the problem can be estimated by training an ensemble of neural networks, thereby generating a distribution on the parameters $\bm{\theta}$ \cite{Gaskin_2024}. This is computationally costly, since a family of neural networks need to be trained in parallel. Meanwhile, in theory the uncertainty on the input data can be `pushed through' the trained neural network, as in \cite{Gaskin_2024c}. Given a prior distribution $\pi_0$ on the input and neglecting the uncertainty on $\bm{\theta}$, the posterior is simply
\begin{equation}
    p(\bm{T}) = ({u_\theta}_\#\pi_0)(\boldsymbol{\chi}),
    \label{eq:UQ}
\end{equation}
with $\#$ indicating a pushforward. For our estimates, we do not know the uncertainty on the inputs except for the initial stock estimate. We train an ensemble of 15 neural networks in parallel to solve the inference problem, and for each draw $n=100$ samples of the initial stocks to estimate overall uncertainty. This gives $1500$ samples of the flow table $\bm{\hat{T}}$. In this article and the accompanying datasets, we provide the average value $\langle \hat{\bm{T}}\rangle$ and one standard deviation.

\subsection{Calculating the elasticity}
The elasticity eq. \eqref{eq:elasticity} is further given by
\begin{equation*}
\nu_i = \left\vert \dfrac{\partial u_\theta}{\partial {x_i}} \right\vert \vert x_i\vert,
\end{equation*}
with $x_i$ the untransformed $i$-th covariate. Since the neural network takes $\psi_\lambda$-transformed covariates as input, we can apply the chain rule to $\chi_i = \psi_\lambda(x_i)$ to obtain
\begin{equation*}
    \nu_i = \left\vert \dfrac{\partial u_\theta}{\partial {\chi_i}} \dfrac{\partial \psi}{\partial {x_i}}\right\vert \vert x_i \vert.
\end{equation*}
\subsection{Calibrating the initial stock value}
To generate a time series of migrant stocks $\hat{S}_{ij}(t)$, an initial condition $S_{ij}(t_0=1990)$ is required to solve the stock evolution equation \eqref{eq:stock_evolution} forward in time. This initial value may be taken from existing sources such as UN DESA where available. However, rather than conditioning on a potentially noisy or inconsistent baseline, we instead calculate a constant offset that best aligns modelled and observed values, while accounting for demographic dynamics. Let $S_{ij}(t)$ be the observed migrant stock, $\hat{S}_{ij}(t)$ the corresponding model prediction generated from an arbitrary initial value $S_{ij}(t_0)$, $w^s_{ij}(t)$ the weight on each observation, and $\gamma_j(t)$ the mortality rate of the destination country $j$. Define the survival fraction $\tilde{\gamma}_j(t)$ as the proportion of individuals alive in 1990 who are still alive in year $t$:
\begin{equation*}
    \tilde{\gamma}_j(t) = \prod_{t_0 < \tau \leq t}(1-\gamma_j(\tau)) \geq 0
\end{equation*}
and $\tilde{\gamma}(t_0)=1$. The optimal offset $b_{ij} \in \mathbb{R}$ of the initial stock value $\hat{S}_{ij}(t_0)$ is then computed by minimizing the weighted squared error between observed and predicted stocks:
\begin{equation}
    b_{ij} = \dfrac{\sum_t \tilde{\gamma}_j w^s_{ij}(S_{ij}(t) - \hat{S}_{ij}(t))}{\sum_t w_{ij}\tilde{\gamma}_j^2} \in \mathbb{R},
\end{equation}
and the baseline-shifted stock then given by $\hat{S}_{ij}(t) +\tilde{\gamma}_j(t)b_{ij}$.
$b_{ij}$ is further constrained to ensure that all resulting stocks remain non-negative.

\appendix
\onecolumn
\cleardoublepage\phantomsection\addcontentsline{toc}{section}{Supporting Information}
\renewcommand\thesection{S\arabic{section}}
\setcounter{section}{0}
\renewcommand\thetable{S\arabic{table}}
\setcounter{table}{0}
\renewcommand\thefigure{S\arabic{figure}}
\setcounter{figure}{0}
{\noindent \huge \bfseries Supporting Information}
\bigskip

\noindent In the following we detail collection and assembly of the input and target data used to train the neural network, as well the validation process used to select the neural network architecture.
\section*{Target data}
\subsection{Net migration data}
\begin{figure*}[ht!]
\begin{adjustbox}{minipage=\textwidth-20pt}
\begin{minipage}{0.5\textwidth}
    \cs{\textbf{I} New Zealand}
\end{minipage}
\begin{minipage}{0.5\textwidth}
    \cs{\textbf{J} Spain}
\end{minipage}
\hfill\vspace{-2mm}

\begin{minipage}{0.5\textwidth}
    \includegraphics[width=\textwidth]{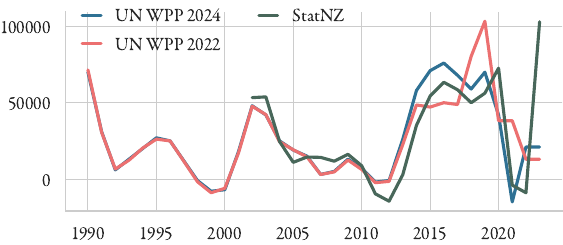}
\end{minipage}
\begin{minipage}{0.5\textwidth}
    \includegraphics[width=\textwidth]{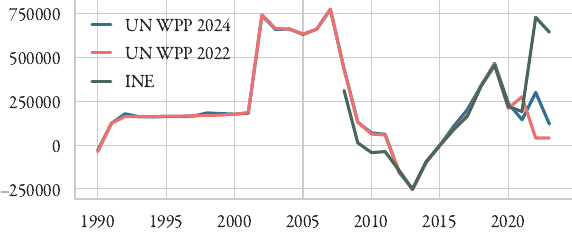}
\end{minipage}
\vspace{0.1em}

\begin{minipage}{0.5\textwidth}
    \cs{\textbf{K} South Korea}
\end{minipage}
\begin{minipage}{0.5\textwidth}
    \cs{\textbf{L} Taiwan}
\end{minipage}
\hfill\vspace{-2mm}

\begin{minipage}{0.5\textwidth}
    \centering
    \includegraphics[width=\textwidth]{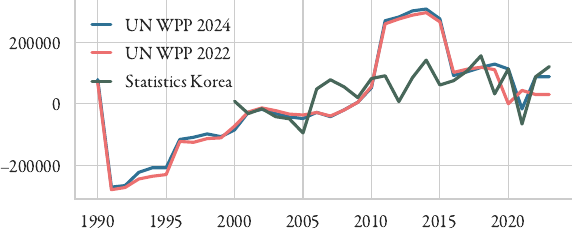}
\end{minipage}
\begin{minipage}{0.5\textwidth}
    \includegraphics[width=\textwidth]{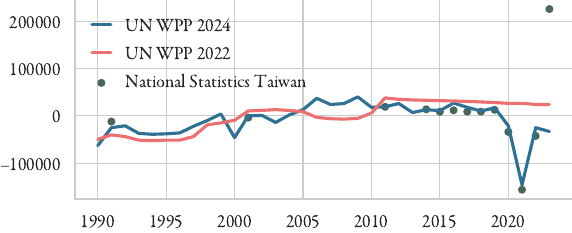}
\end{minipage}
\vspace{0.1em}

\begin{minipage}{\textwidth}
    \cs{\textbf{M} United States}
\end{minipage}
\hfill\vspace{-2mm}

\begin{minipage}{\textwidth}
    \centering
    \includegraphics[width=\textwidth]{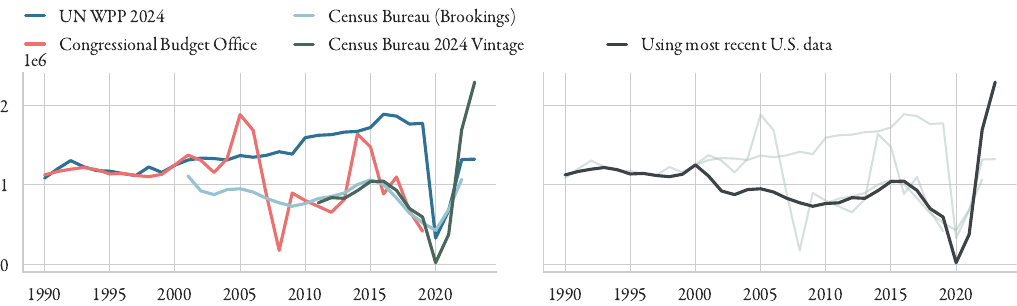}
\end{minipage}
\caption{Selected net migration statistics from national statistical offices alongside UN WPP 2024 and UN WPP 2022 figures.}
\label{fig:SI:net_migration}
\end{adjustbox}
\end{figure*}

To train the neural network, we use target net migration statistics for a small number of countries. This data are mostly taken from National Statistics Bureaus, as well as the UN 2024 World Population Prospects for a handful of additional countries and periods. In figures \ref{fig:SI:net_migration} we show national figures alongside the UN WPP 2024 and WPP 2022 Revision data for some selected countries. UN WPP 2024 figures generally agree well with national estimates for periods pre-2010; larger discrepancies typically occur around 2020 due to the pandemic, the war in Ukraine, and a probable lack or lag in new demographic data reaching demographers at UN DESA.
\begin{figure*}[ht!]
\begin{adjustbox}{minipage=\textwidth-20pt}
\begin{minipage}{0.5\textwidth}
    \cs{\textbf{A} Australia}
\end{minipage}
\begin{minipage}{0.5\textwidth}
    \cs{\textbf{B} Canada}
\end{minipage}
\hfill\vspace{-2mm}

\begin{minipage}{0.5\textwidth}
    \includegraphics[width=\textwidth]{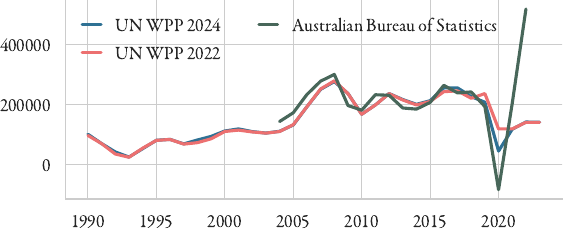}
\end{minipage}
\begin{minipage}{0.5\textwidth}
    \includegraphics[width=\textwidth]{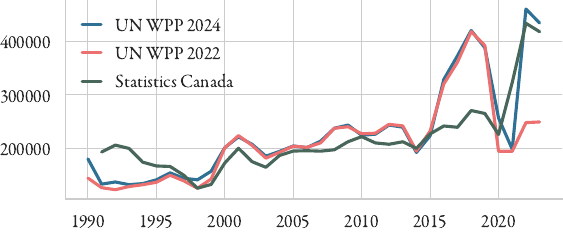}
\end{minipage}
\vspace{0.1em}

\begin{minipage}{0.5\textwidth}
    \cs{\textbf{C} Germany}
\end{minipage}
\begin{minipage}{0.5\textwidth}
    \cs{\textbf{D} Finland}
\end{minipage}
\hfill\vspace{-2mm}

\begin{minipage}{0.5\textwidth}
    \includegraphics[width=\textwidth]{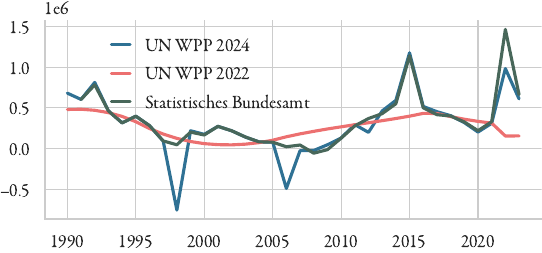}
\end{minipage}
\begin{minipage}{0.5\textwidth}
    \includegraphics[width=\textwidth]{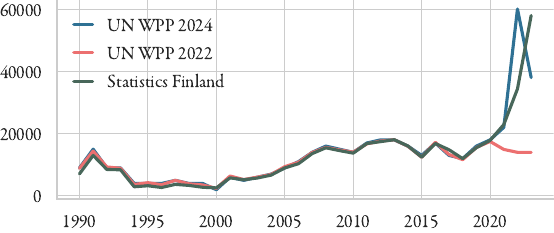}
\end{minipage}
\vspace{0.1em}

\begin{minipage}{0.5\textwidth}
    \cs{\textbf{E} France}
\end{minipage}
\begin{minipage}{0.5\textwidth}
    \cs{\textbf{F} Ireland}
\end{minipage}
\hfill\vspace{-2mm}

\begin{minipage}{0.5\textwidth}
    \includegraphics[width=\textwidth]{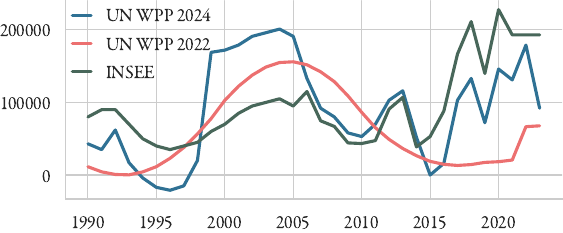}
\end{minipage}
\begin{minipage}{0.5\textwidth}
    \includegraphics[width=\textwidth]{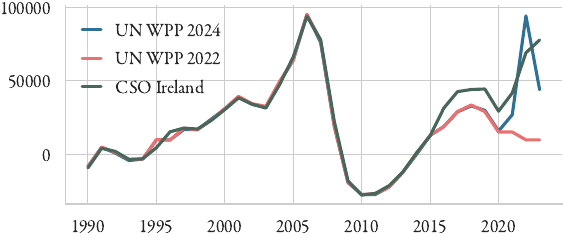}
\end{minipage}
\vspace{0.1em}

\begin{minipage}{0.5\textwidth}
    \cs{\textbf{G} Italy}
\end{minipage}
\begin{minipage}{0.5\textwidth}
    \cs{\textbf{H} Japan}
\end{minipage}
\hfill\vspace{-2mm}

\begin{minipage}{0.5\textwidth}
    \includegraphics[width=\textwidth]{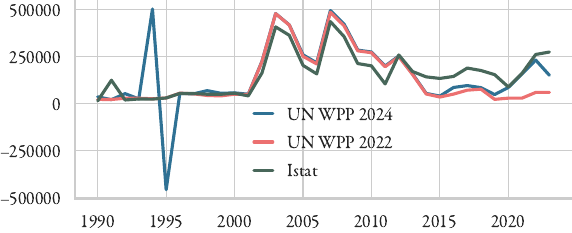}
\end{minipage}
\begin{minipage}{0.5\textwidth}
    \includegraphics[width=\textwidth]{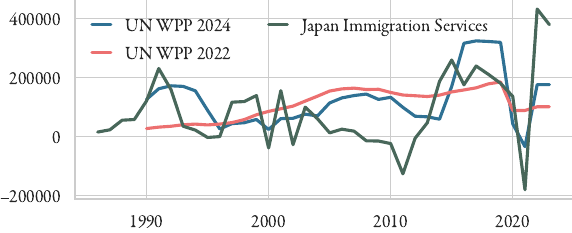}
\end{minipage}
\end{adjustbox}
\end{figure*}
\begin{itemize}
    \item Australia: data from 2004 onwards provided by the Australian Bureau of Statistics \cite{ABS_data}. Pre-2004 we use UN WPP 2024 data.
    \item Austria: data from 2002 onwards provided by Statistik Austria \cite{Statistik_Austria}. Pre-2002 we use UN WPP 2024 data.
    \item Belgium: data from 2000 onwards provided by StatBEL \cite{StatBEL}. Pre-2000 we use UN WPP 2024 data.
    \item Bulgaria: data from 2010 onwards provided by the National Statistical Office of Bulgaria \cite{Statistics_Bulgaria}.
    \item Canada: data provided by Statistics Canada \cite{Statistics_Canada}. Since figures are given for years ending in the summer, we use the midpoint average as an approximation for calendar-year net migration figures.
    \item Czech Republic: data provided by the Czech Statistical Office \cite{Statistics_Czechia}.
    \item Denmark: data provided by Statistics Denmark \cite{Stat_Denmark_net_migration}.
    \item Estonia: data from 2000 onwards provided by Statistics Estonia \cite{Stat_Estonia}; since the figures agree with the UN WPP 2024 data, pre-2000 we use UN WPP 2024 data.
    \item Finland: data provided by Statistics Finland \cite{Stat_Finland_net_migration}.
    \item France: data are provided by INSEE (Institut National de la statistique et des \'{e}tudes \'{e}conomiques) \cite{INSEE_data}.
    \item Germany: data are provided by the Statistisches Bundesamt \cite{Statistisches_Bundesamt}.
    \item Iceland: data provided by Statistics Iceland \cite{Stat_Iceland_net_migration}.
    \item Ireland: data provided by the Central Statistics Office \cite{CSO_IRL_data}; these are counted from April to April and thus need to be shifted back and interpolated accordingly.
    \item Italy: data from Istat \cite{Istat_historical, Istat}.
    \item Japan: data are provided by the Japan Immigration Services Agency \cite{Statistics_Japan}.
    \item Latvia: data are provided by the Statistics Agency of Latvia \cite{Statistics_Latvia}.
    \item Lithuania: data are provided by Statistics Lithuania \cite{Statistics_Lithuania}.
    \item Netherlands: data from 2003 onwards is provided by Statistics Netherlands \cite{StatNL} agree with the UN WPP 2024 figures, and these are thus used for the remaining years.
    \item New Zealand: data from 2002 onwards is provided by Statistics New Zealand \cite{Stat_NZ_net_migration}. Pre-2002 we use UN WPP data.
    \item Norway: data are provided by Statistics Norway \cite{Stat_Norway_net_migration}.
    \item Portugal: data from 2002 onwards provided by Statistics Portugal \cite{Statistics_Portugal}. Pre-2002 we use UN WPP 2024 data.
    \item Slovenia: data are provided by the Statistical Office of Slovenia \cite{Statistics_Slovenia}.
    \item South Korea: Statistics Korea \cite{Statistics_Korea} publishes net migration figures from 2000 onwards. Pre-2000 values are not used.
    \item Spain: data from 2008 onwards is provided by INE (Institudo Nacional de Estad\'{i}stica) \cite{INE_data}. Pre-2008 we use UN WPP 2024 data.
    \item Sweden: data provided by Statistics Sweden \cite{Stat_Sweden_net_migration}.
    \item Switzerland: data are provided by the Bundesamt f\"{u}r Statistik \cite{Bundesamt_Statistik_CH} from 1991; for 1990, we use the UN WPP 2024 figure.
    \item Taiwan: data for some years is provided by National Statistics Taiwan \cite{Statistics_TWN}. Missing years are filled using the UN WPP 2024 data.
    \item United Kingdom: data are provided by the Office for National Statistics \cite{ONS_net_migration}.
    \item United States: figures from the US Census Bureau \cite{Brookings_2023,USCB_Net_migration}, the Congressional Budget Office \cite{CBO_net_migration}, and the UN WPPs offer inconsistent pictures (see fig. \ref{fig:SI:net_migration}{\cs{\textbf{M}}}). We collate the data by always using the most recent figures from US bureaus, shown on the right.
\end{itemize}

\subsection{Migrant stocks}
\paragraph{Interpolation and Extrapolation} The stocks are both a target value and an input to the neural network. We use the UN DESA 2025 Revision data where available, and fill gaps using values from the 2020 Revision. For edges $(i, j, k)$ where stocks $S_{ij}(t)$ or $S_{ik}(t)$ are missing for \emph{all} time points $t$, we set the flow $T_{ijk}(t) = T_{ikj}(t) = 0$ for all $t$ if $i \neq j$ or $i \neq k$, and set the initial stock value to 0. This is problematic as it underestimates some fairly major edges (e.g. Argentina or Brazil to India), but there is no stock data available from which an inference could be made. However, there are many series $S_{ij}(t)$ where data are partially observed (i.e. values are missing for some, but not all time points $t$). For these series, we extrapolate missing stocks in the following way: let $c_{jk} = \mathrm{corr}(S_{ij}, S_{ik})$ and $\rho_{jk} = \exp(-d_{jk})$, where $d_{jk}$ is the geodesic distance between country $j$ and $k$.

\begin{figure*}[t!]
\begin{adjustbox}{minipage=\textwidth-20pt}
\begin{minipage}[t]{\textwidth}
    \includegraphics[width=\textwidth]{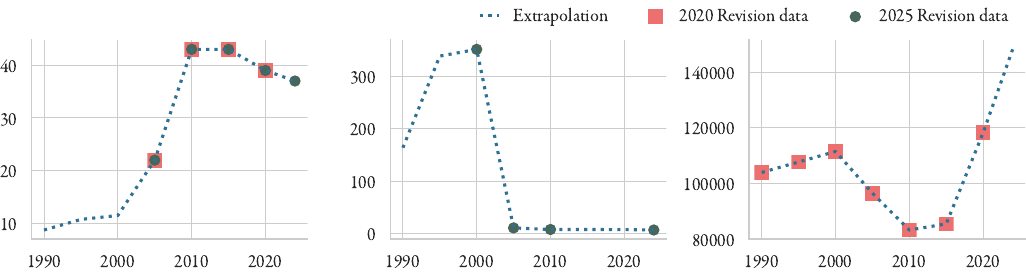}
\end{minipage}
\caption{Three examples of stock data interpolation and extrapolation, in cases where we extrapolate back (left), forwards (right), and also interpolate intermediate values (centre). Stocks are interpolated using the weighted average growth rates of similar countries, see text. Shown are the original mid-year estimates.}
\label{fig:SI:stock_interpolation}
\end{adjustbox}
\end{figure*}
We want to extrapolate missing stock values by comparing them to `similar' countries with complete data. `Similarity' here means that (1) the stock time series correlate strongly, and (2) the countries are geographically close. We thus build a weight measure for a country pair $(j, k)$ using
\begin{equation*}
    w_{jk} = c_{jk} \times \dfrac{\rho_{jk}}{\sum_k \rho_{jk}}.
\end{equation*}
Then, given a matrix of stock growth values $S_{ij} / S_{ij}(t-1) - 1$, we define an average growth rate for the missing series as 
\begin{equation*}
    \bar{g}_{ij} = \dfrac{\sum_kg_{ik}w_{jk}}{\sum_k w_{jk}}.
\end{equation*}
The missing stocks are then extrapolated and interpolated using this growth rate. This can be done even in the case of only a single datapoint (see fig. \ref{fig:SI:stock_interpolation}).

After interpolation, the native-born population $S_{ii}$ is then estimated as 
\begin{equation*}
    S_{ii}(t) = P_i(t) - \sum_{j \neq i} S_{ji},
\end{equation*}
where $P_i(t)$ is the total population (in July of each year), taken from the UN WPP dataset. The native-born population naturally makes up the upper tail of the migrant stock distribution (see fig. \ref{fig:SI:Covariate_scaling}{\cs{\textbf{H}}}).

\paragraph{Estimating beginning-of-year stocks} The UN DESA stocks are mid-year values, but in order to align the data with the net migration and flow values, we need to estimate the stocks at the beginning of each year. We do this using iterative procedural fitting: let $\bm{S}$ be a stock table at mid-year, $\bm{B}(t)$ the vector of total births in year $t$, and $\bm{\gamma}(t)$ the annual death rate in year $t$. We then scale $\bm{S}$ to match the row marginal 
\begin{equation*}
    \sum_i \dfrac{S_{ij}}{\sqrt{1-\gamma_j}} - \dfrac{B_i}{2\sqrt{1-\gamma_i}},
\end{equation*}
where the factor of $2$ is a result of assuming that births are distributed evenly throughout the year, i.e. half the births occur in the first six months of each year. The column marginals represent the total population in January of each year, as given by the UN WPP dataset \cite{UN_WPP_2024}.

\begin{figure*}[t!]
\begin{adjustbox}{minipage=\textwidth-20pt}
\begin{minipage}[t]{\textwidth}
    \includegraphics[width=\textwidth]{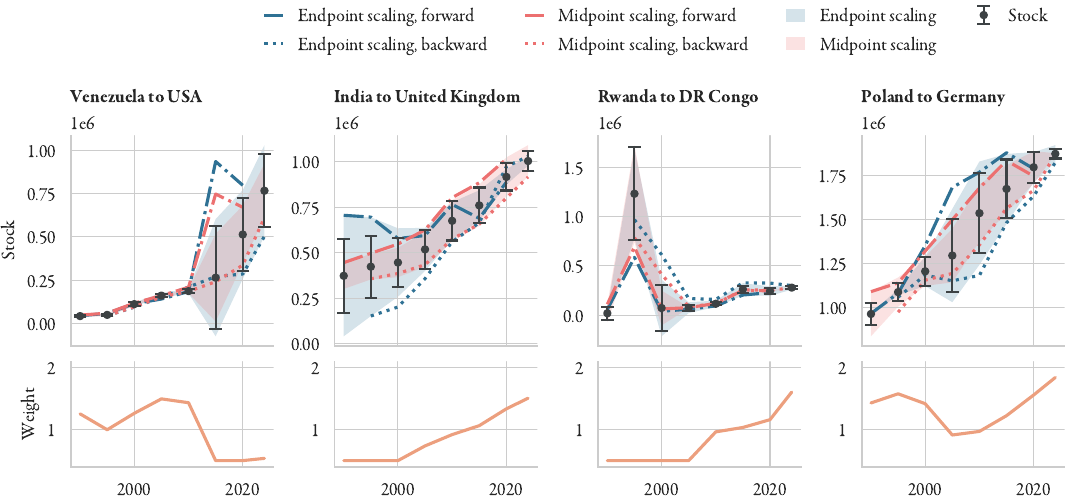}
\end{minipage}
\caption{Calculating the weights for the stock target values. Using demographic accounting and iterative proportional fitting, between two and four new stock values can be obtained for each year (`endpoint scaling' and `midpoint scaling'). Shaded areas indicate the average distance for each type of scaling (midpoint vs. endpoint), and the errorbars show the UN DESA value with the resulting overall error. The average distance to the UN DESA value (at the start of year) is then used to calculate a weight (eq. \eqref{eq:stock_weight}, bottom row) used in the loss function during training.}
\label{fig:SI:stock_weights}
\end{adjustbox}
\end{figure*}
\begin{figure*}[ht!]
\begin{adjustbox}{minipage=\textwidth-20pt}
\begin{minipage}[t]{\textwidth}
    \includegraphics[width=\textwidth]{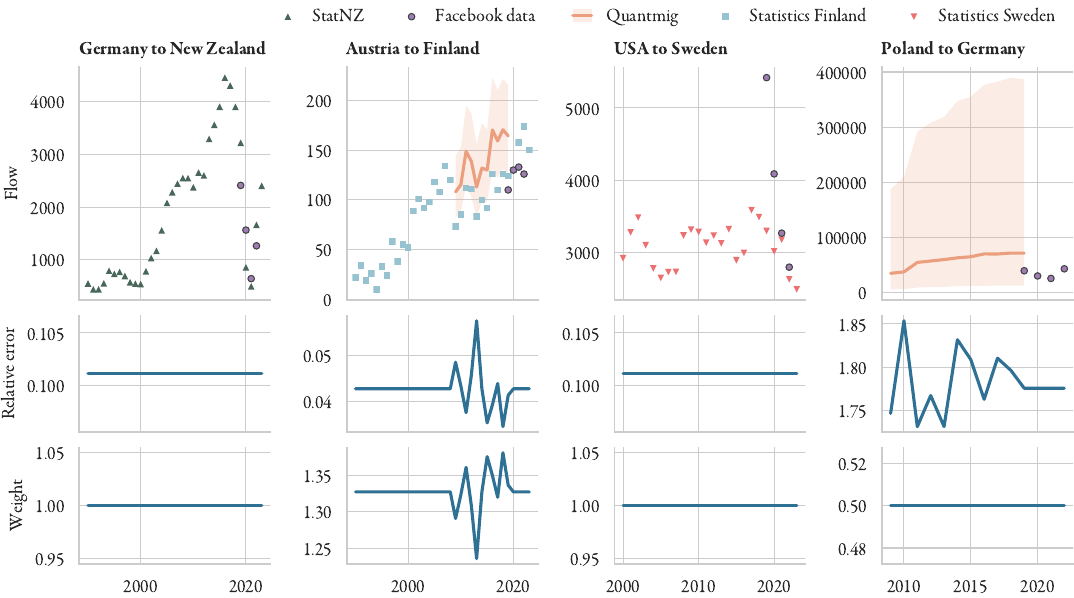}
\end{minipage}
\caption{Calculating the weights for the flow targets. Where given, the weights are calculated from the relative error on the flow data---this is only the case for the Quantmig dataset (orange). Where uncertainty is at least partially observed, we set the weight on the remainder of the flow corridor to the median weight of the Quantmig data points (see e.g. the Austria to Finland corridor). Where the uncertainty on a flow corridor is wholly unobserved (e.g. Germany to New Zealand), the weight is set to 1. The weights are clipped to the interval $[0.5, 2]$ to ensure that no data are entirely lost or overly dominates the loss function (see e.g. the Poland to Germany corridor, which comes with significant uncertainty).}
\label{fig:SI:flow_weights}
\end{adjustbox}
\end{figure*}
\begin{figure*}[t!]
\begin{adjustbox}{minipage=\textwidth-20pt}
\begin{minipage}{0.33\textwidth}
    \cs{\textbf{A} Weights on stocks}    
\end{minipage}
\begin{minipage}{0.33\textwidth}
    \cs{\textbf{B} Weights on flows}    
\end{minipage}
\begin{minipage}{0.33\textwidth}
    \cs{\textbf{C} Weights on net migration}    
\end{minipage}
\hfill\vspace{-2mm}

\begin{minipage}[t]{\textwidth}
    \includegraphics[width=\textwidth]{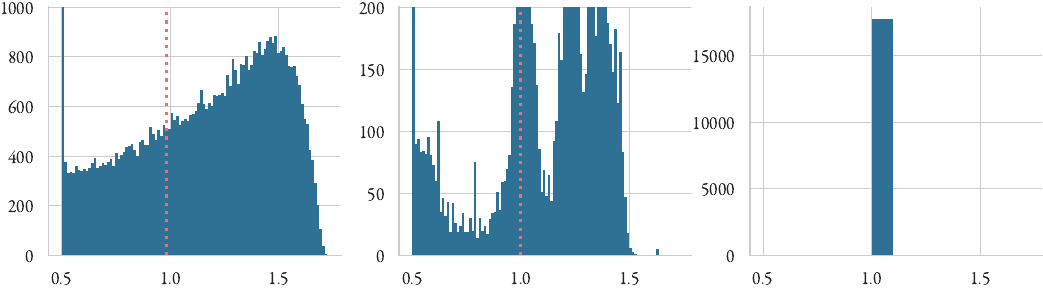}
\end{minipage}
\caption{Distribution of the weights on the stock, flow, and net migration target values. The distributions are clipped to $[0.5, 2]$, and each have mean 1 (red dotted lines). For the net migration, we currently set all weights to 1.}
\label{fig:SI:weight_distribution}
\end{adjustbox}
\end{figure*}

\paragraph{Calculating the weights} We generate demographically closed stock matrices using iterative proportional fitting. Let $\bm{S}_1 = \bm{S}(t_1)$, $\bm{S}_2 = \bm{S}(t_2)$ be two successive stock tables, i.e. $t_2 = t_1 +5$. In principle, adding the total number of births in the period $[t_1, t_2]$ to $\bm{S}_1$ and the total deaths in $[t_1, t_2]$ to $\bm{S}_2$ should result in stock matrices with the same marginals $a=\sum_i S_{ij}$, $b=\sum_j S_{ij}$. In practice, this is not the case. We can thus calculate the marginals $a_1$, $b_1$, $a_2$, $b_2$ for each stock matrix with births and deaths added, and scale each to the mid-point $\frac{1}{2}(a_1 + a_2)$, $\frac{1}{2}(b_1 + b_2)$ using IPF (`midpoint scaling'). After scaling, we subtract births and deaths again, obtaining new estimates of the stock table for each year. Doing this for each pair of stock tables, we obtain two estimates of the stock for each year except the years at the boundaries (1990 and 2024), for which we only have one; one value comes from comparison with the previous stock table (`backward') and one from comparison with the next table (`forward'). Additionally, we can also add births \emph{and} subtract deaths \emph{only} from $S_1$, and repeat the same procedure, obtaining two more estimates (`endpoint scaling'). In total, we thus obtain four additional estimates for each year except the boundary years, for which we have two (see fig. \ref{fig:SI:stock_weights}). The error on each stock value is then given by 
\begin{equation}
    \sigma_{ij} = \langle \vert S_{ij}^k - S_{ij}\vert \rangle_k,
\end{equation}
where $\langle \cdot \rangle_k$ is the average over the various estimates obtained from IPF. The weights are calculated based on the relative error $\rho_{ij} = \sigma_{ij} / S_{ij}$, where we first normalise $\rho_{ij}$ to have mean 0 and variance 1; this ensures that the weights are balanced around 1. We also want to avoid excessive distortion of the loss function due to extremely large or extremely small weights, and hence truncate the weights to $[0.5, 2]$. The weights are thus given by
\begin{equation}
    w_{ij} = \min\left(2,\max\left(0.5, \exp(-(\rho_{ij} - \langle \rho_{ij} \rangle)/ \langle \rho_{ij}^2 \rangle^{1/2})\right)\right),
    \label{eq:stock_weight}
\end{equation}
see fig. \ref{fig:SI:weight_distribution}{\cs{\textbf{A}}}. We estimate the error on a stock difference $S_{ij}(t_2) - S_{ij}(t_2)$ by assuming the errors are independent, setting $\sigma(\Delta S) = \sqrt{\sigma_1^2 + \sigma_2^2}$, and calculating the weights as above.

\subsection{Flow data}
The uncertainty on the flow targets is based on the uncertainty from the Quantmig estimates, which come with standard errors. As with the stocks, the weights are then given by 
\begin{equation}
    w_{ij} = \min\left(2,\max\left(0.5, \exp(-(\rho_{ij} - \langle \rho_{ij} \rangle)/ \langle \rho_{ij}^2 \rangle^{1/2})\right)\right),
\end{equation}
with $\rho_{ij}$ the relative error (standard error divided by central estimate). For all flow corridors along which we partially know the median errors, we set the relative error to the median relative error on that corridor, and calculate the weights as above. Finally, for flow corridors with entirely unobserved errors, we set the relative error to the mean of the Quantmig dataset and the weight to 1 (see fig. \ref{fig:SI:flow_weights}).

\section*{Input data}
\subsection{GDP per capita and GDP annual growth}
In this section we detail the assembly of the economic covariates used to train the neural network: GDP per capita, in real 2015 USD; and GDP \% annual growth, given by
\begin{equation}
    \partial_t \text{GDP}(t) = 100 \times \left(\dfrac{\text{GDP}(t)}{\text{GDP}(t-1)} - 1\right).
    \label{eq:GDP_growth}
\end{equation}
GDP data are sourced from the World Bank \cite{WB_GDP_growth,WB_GDP_PCAP} as well as UNCTAD \cite{UNCTAD_GDP}. However, these datasets are not complete for all countries and years required. Historical GDP growth rates can be calculated from the Maddison dataset \cite{Maddison_GDP} (giving GDP/capita in 2011 PPP) and the IMF dataset \cite{IMF_WEO_2024} (GDP/capita in 2021 PPP). Note that GDP growth can be calculated from any measure of real GDP (e.g. constant USD or constant PPP). GDP growth, when available, can be used to calculate GDP/capita for missing years. At times, only nominal GDP figures are available for a given year and country. Converting to real GDP requires the \emph{deflator} for that country, 
\begin{equation}
    \text{GDP deflator}(t) = \dfrac{\text{nominal GDP}(t)}{\text{real GDP}(t)}.
\end{equation}
Where the deflator is unavailable, deflators from economically, geographically, or socially similar countries can be used to estimate real GDP. We also fill missing values using figures from national statistical bureaus and additional sources, as below. 

GDP data are scaled using a Yeo-Johnson transform (see fig. \ref{fig:SI:Covariate_scaling}{\cs{\textbf{A}}}--{\cs{\textbf{B}}}). 

\subsubsection{Caribbean Netherlands}
\begin{figure*}[t!]
\begin{adjustbox}{minipage=\textwidth-20pt}
\begin{minipage}{0.53\textwidth}
    \cs{\textbf{A} Nominal GDP per capita}
\end{minipage}
\begin{minipage}{0.47\textwidth}
    \cs{\textbf{B} Real GDP per capita, in const. 2015 USD}
\end{minipage}
\hfill\vspace{-2mm}

\begin{minipage}{\textwidth}
    \centering
    \includegraphics[width=\textwidth]{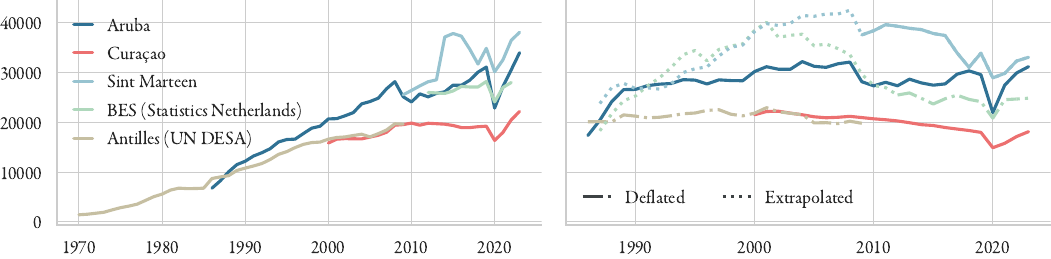}
\end{minipage}
\caption{Caribbean Netherlands. \textbf{A} Available nominal GDP per capita figures for the constituent islands. \textbf{B} We fill missing years by deflating nominal GDP figures using the deflator for Aruba, and then extrapolate back using the growth rate from Aruba.}
\label{fig:SI:Antilles}
\vspace{5mm}
\begin{minipage}[ht!]{0.49\textwidth}
    \centering
    \includegraphics[width=\textwidth]{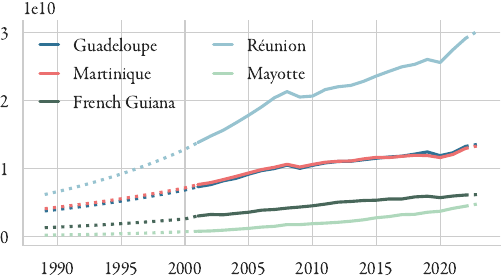}
    \caption{French overseas territories real GDP, in 2015 USD. The available data from Statista is extrapolated to the entire period 1989--2023 using the average growth rates for each country.}
\label{fig:SI:FR_overseas}
\end{minipage}
\hspace{0.02\textwidth}
\begin{minipage}[ht!]{0.49\textwidth}
    \centering
    \includegraphics[width=\textwidth]{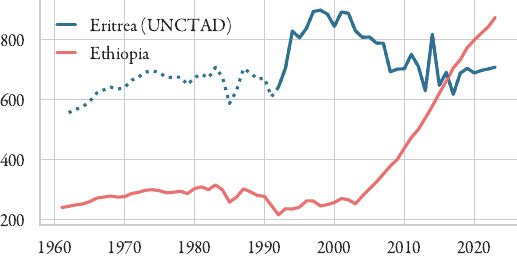}
    \caption{Eritrea. Since the country did not formerly achieve independence from Ethiopia until 1991, we extrapolate the available data from UNCTAD back to 1989 using the Ethiopian growth rate (dotted line).}
\label{fig:SI:Eritrea}
\end{minipage}
\end{adjustbox}
\end{figure*}

The Netherlands Antilles were a constituent country of the Kingdom of the Netherlands, consisting of the islands of Bonaire, St. Eustatius, and Saba (the BES islands), Aruba, Cura\c{c}ao, and Sint Marteen. Aruba became an independent constituent country in 1986, and in 2010, Cura\c{c}ao and Sint Marteen followed suit, at which point the Netherlands Antilles were dissolved and the remaining BES islands became a special municipality within the Kingdom. The World Bank and UNCTAD provide real GDP figures for Aruba from 1986, for Curacao from 2000, and for Sint Marteen from 2009. Statistics Netherlands \cite{BES_GDP} provides nominal GDP for the BES islands from 2012--2022, while UN DESA provides nominal GDP for the entire Netherlands Antilles from 1970--2009 \cite{UN_DESA_data}, which agrees with the World Bank data for Cura\c{c}ao, the largest of the remaining Antilles islands after 1986 (see fig. \ref{fig:SI:Antilles}{\cs{\textbf{A}}}). Since the deflators are not available, we convert the UN DESA data to real GDP using the deflator for Aruba, and then extrapolate back to 1986 using the growth rate for Aruba (see fig. \ref{fig:SI:Antilles}{\cs{\textbf{B}}}).

\subsubsection{French overseas territories}
Statista \cite{Statista_FR_overseas_GDP} provides real GDP in constant 2014 Euros for the period 2000--2022. We convert to 2015 USD and extrapolate back to 1989 using the average growth rate for each territory, calculated over the period from 2000--2008. We extrapolate forward to 2023 using the average growth rate for each country in the period 2017--2022 (see fig. \ref{fig:SI:FR_overseas}).
\subsubsection{Eritrea}
Eritrea formally achieved independence from Ethiopia in 1991. UNCTAD provides data from 1992, and we extrapolate back to 1989 using the growth rate from Ethiopia (see fig. \ref{fig:SI:Eritrea}).
\subsubsection{Small Pacific Islands}
\begin{figure*}[t!]
\begin{adjustbox}{minipage=\textwidth-20pt}
\begin{minipage}{0.49\textwidth}
    \centering
    \includegraphics[width=\textwidth]{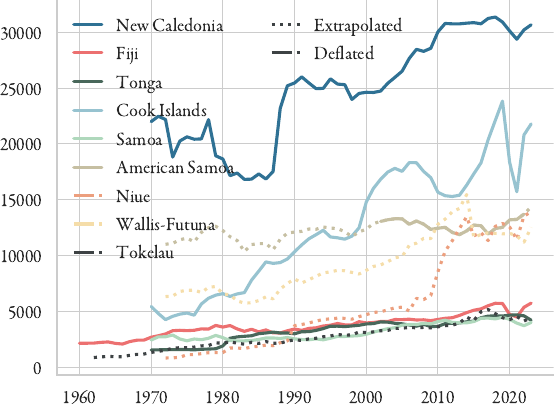}
    \caption{Small Pacific Islands GDP per capita, in const 2015 USD. Nominal GDP data from the Pacific Community data are deflated using the inflation rate as a proxy (dash-dotted lines). We then extrapolate back (dotted lines) using growth rates from neighbouring countries with similar GDP/capita.}
\label{fig:SI:SPI}
\end{minipage}
\hspace{0.02\textwidth}
\begin{minipage}{0.49\textwidth}
    \centering
    \includegraphics[width=\textwidth]{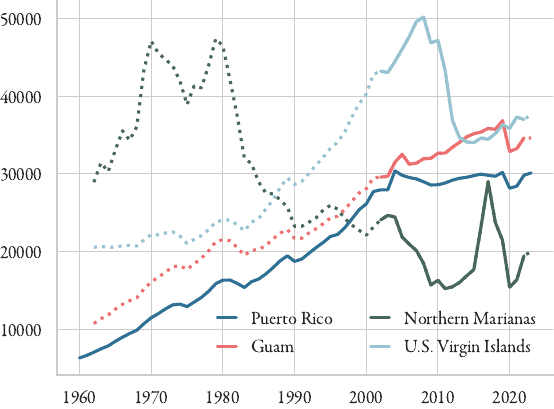}
    \caption{US overseas territories GDP per capita, in const 2015 USD. Missing values for Guam, the Northern Marianas, and the US Virgin Islands are extrapolated back using the growth rate of Puerto Rico (dotted lines).}
\label{fig:SI:US_overseas_territories}
\end{minipage}

\vspace{5mm}
\begin{minipage}{0.49\textwidth}
    \centering
    \includegraphics[width=\textwidth]{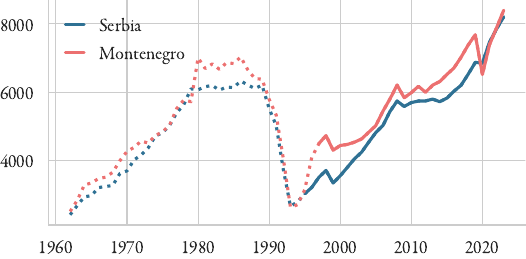}
    \caption{Serbia and Montenegro, GDP per capita in const 2015 USD. We extrapolate back to 1960 using the growth rates given by the Maddison project dataset (dotted lines).}
\label{fig:SI:SRB_MNE}
\end{minipage}
\hspace{0.02\textwidth}
\begin{minipage}{0.49\textwidth}
    \centering
    \includegraphics[width=\textwidth]{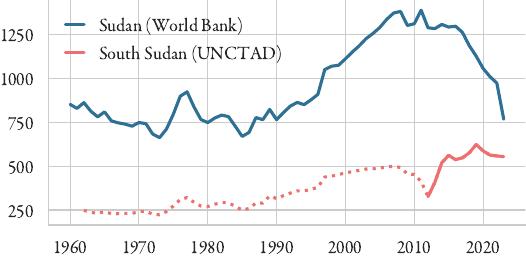}
    \caption{Sudan and South Sudan GDP per capita in const 2015 USD. We extrapolate the UNCTAD data for South Sudan back from 2011, the year of its independence, using the growth rate of Sudan (dotted line).}
\label{fig:SI:SSD}
\end{minipage}
\end{adjustbox}
\end{figure*}
The World Bank and UNCTAD datasets have missing data for American Samoa, Niue, Tokelau, Wallis and Futuna, Guam, and the Northern Mariana Islands. The Pacific Community provides nominal GDP figures \cite{Pacific_Community_GDP}, as well as country-specific inflation rates \cite{Pacific_Community_inflation_rate}, which we use as a proxy for the GDP deflator. We then extrapolate back using growth rates of neighbouring countries:
\begin{enumerate}
    \item American Samoa, Niue, and Wallis-Futuna: average growth rate of New Caledonia, the Cook Islands, and Fiji,
    \item Tokelau: growth rate of Fiji.
\end{enumerate}

\begin{figure*}[t!]
\begin{adjustbox}{minipage=\textwidth-20pt}
\begin{minipage}[t]{0.49\textwidth}
    \centering
    \includegraphics[width=\textwidth]{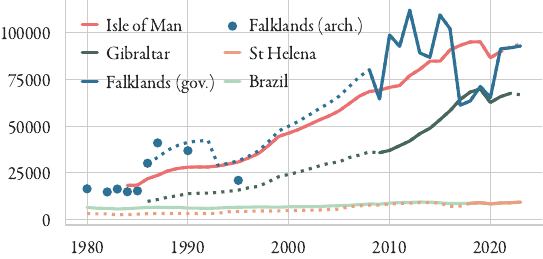}
    \caption{British territories real GDP per capita in const 2015 USD. Missing data for Gibraltar and Falklands is extrapolated back from Government data using the Isle of Man growth rate (dotted lines). The Falklands extrapolation agrees broadly with archival data (blue dots). Government data for St. Helena is extrapolated back using the growth rate for Brazil, also shown for comparison purposes.}
\label{fig:SI:UK_overseas}
\end{minipage}
\hspace{0.02\textwidth}
\begin{minipage}[t]{0.49\textwidth}
    \centering
    \includegraphics[width=\textwidth]{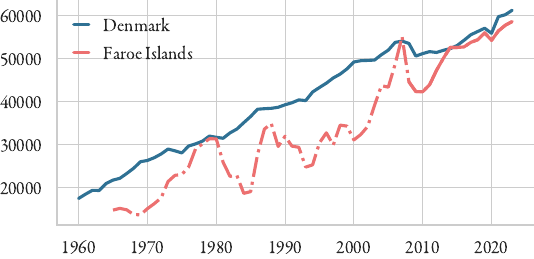}
    \caption{Faroe Islands real GDP per capita in const 2015 USD. Missing data for the islands is calculated by deflating the available nominal figures using the Danish deflator as a proxy (dash-dotted line).}
\label{fig:SI:FRO}
\end{minipage}
\end{adjustbox}
\end{figure*}

\subsubsection{US overseas territories}
Guam and the Northern Marianas are US overseas territories that to a large part rely on US funding for their economies. We extrapolate both Guam and the Marianas using the growth rate from Puerto Rico. The US Virgin Islands we extrapolate back using the average growth rate of Puerto Rico, St. Kitts and Nevis, and Antigua and Barbuda.

\subsubsection{Serbia and Montenegro}
The state of Serbia and Montenegro existed from 1992 until 2006, at which point the two constituent republics separated and became independent nations. Growth rates for these two countries are estimated in the Maddison project dataset, and are used to extrapolate GDP figures back to 1960 (see fig. \ref{fig:SI:SRB_MNE}).

\subsubsection{South Sudan}
South Sudan ceded from Sudan in 2011. UNCTAD gives real GDP figures from 2011 onwards, and we extrapolate back in time using the growth figures from Sudan (see fig. \ref{fig:SI:SSD}).

\subsubsection{East Timor}
The only missing data point is the GDP growth rate for 1990. A report by the Australian government \cite{East_Timor_GDP_growth} suggests an average 10\% economic growth rate from 1987--1997, due to heavy investment in construction. Given the data we already have, this implies GDP growth from 1987-1990 must have averaged 11.7\%.

\subsubsection{Western Sahara, Saint Pierre Miquelon}
No data are available for these territories. Western Sahara is under Moroccan control, and Saint Pierre Miquelon is a small French territory located near the Canadian province of Newfoundland. We thus simply use Moroccan and Canadian GDP per capita figures respectively.

\subsubsection{British overseas territories and Crown Dependencies}
Four British territories have missing data: the Isle of Man, Gibralatar, the Falkland Islands, and Saint Helena, Ascension and Tristan da Cunha (in the following simply referred to as St. Helena), a small island territory in the Atlantic ocean. The Isle of Man government \cite{IMN_GDP} provides the missing real GDP growth figures for 2022 and 2023, from which we can infer the missing real GDP/capita figures for those years. The government of Gibraltar \cite{GIB_GDP} provides nominal data from 2009; we deflate using the British deflator as a proxy, and extrapolate back using the Isle of Man growth rate. Data for the Falklands real GDP is available from the government website, given at constant 2012 Falklands pounds (pegged to the British pound at an exchange rate of 1:1) \cite{FLK_gov_GDP}. We extrapolate back using the Isle of Man growth rates. Some historical data from 1980--1995 is available from the archives \cite{FLK_archive}, which, when converted to real 2015 USD, agrees well with this extrapolation (see fig. \ref{fig:SI:UK_overseas}). Lastly, nominal GDP data are provided by the St. Helena government website up to 2017. We use the British deflator to convert to real GDP and extrapolate back using the GDP growth rate for Brazil, a nearest neighbour with a similar GDP per capita.
\subsubsection{Faroe Islands}
The World Bank provides real GDP/capita from 2008 onwards, and nominal GDP/capita from 1965 onwards, which we deflate using the Danish deflator to obtain a complete time series, see fig. \ref{fig:SI:FRO}.

\subsection{Bilateral trade}
\begin{figure*}[t!]
\begin{adjustbox}{minipage=\textwidth-20pt}
    \centering
\begin{minipage}{0.5\textwidth}
    \cs{\textbf{A} Total trade flow, Poland to Russia}
\end{minipage}
\begin{minipage}{0.49\textwidth}
    \cs{\textbf{B} Total trade flow, Germany to France}
\end{minipage}
\hfill\vspace{2mm}

\includegraphics[width=\textwidth]{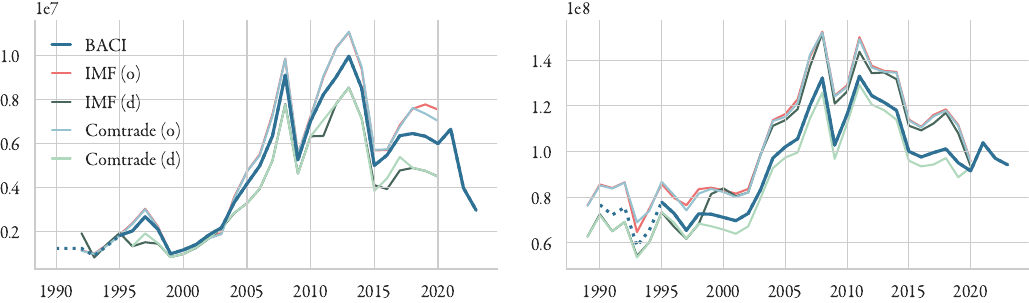}
\caption{Two tradeflow examples, showing total flow in constant 2015 USD. Shown are the BACI figures (solid blue line), as well as the Comtrade and IMF figures. Missing figures in the BACI dataset are extrapolated back using average growth rates (dotted lines).}
\label{fig:SI:trade}
\end{adjustbox}
\end{figure*}

Trade flows are taken from the BACI harmonised tradeflow dataset \cite{BACI,BACI_dataset}, and extrapolated back using the real growth rates calculated from the UN Comtrade and IMF Direction of trade statistics as given in the CEPII gravity dataset \cite{CEPII}. Since each dataset contains both origin- and destination-reported trade flows, we take the average growth rate (see fig. \ref{fig:SI:trade}). Where values are missing, we simply set the trade flow to the last known value. If all values are NaN, we set the flow to $0$.

\begin{figure*}[t!]
\begin{adjustbox}{minipage=\textwidth-20pt}
\begin{minipage}[t]{0.5\textwidth}
    \cs{\textbf{A} GDP per capita, const 2015 USD}
\end{minipage}
\begin{minipage}[t]{0.5\textwidth}
    \cs{\textbf{B} GDP annual growth (\%)}
\end{minipage}
\hfill\vspace{-2mm}

\begin{minipage}[t]{0.5\textwidth}
    \includegraphics[width=\textwidth]{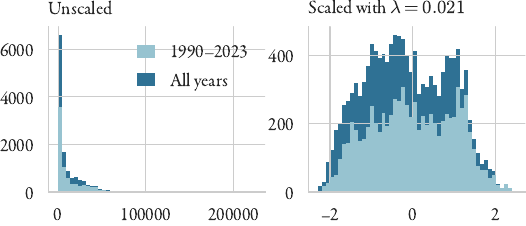}
\end{minipage}
\begin{minipage}[t]{0.5\textwidth}
    \includegraphics[width=\textwidth]{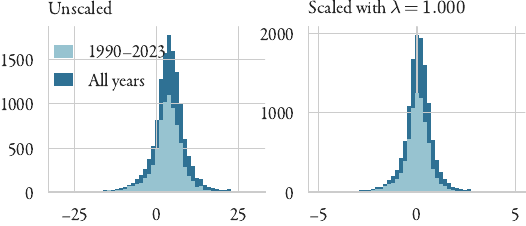}
\end{minipage}

\begin{minipage}[t]{0.5\textwidth}
    \cs{\textbf{C} Total population}
\end{minipage}
\begin{minipage}[t]{0.5\textwidth}
    \cs{\textbf{D} Life expectancy}
\end{minipage}
\hfill\vspace{-2mm}

\begin{minipage}[t]{0.5\textwidth}
    \includegraphics[width=\textwidth]{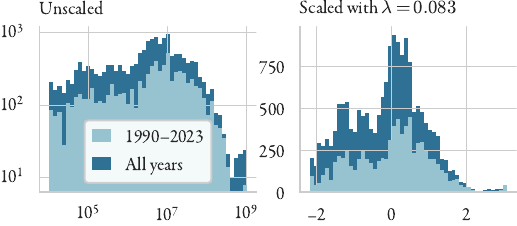}
\end{minipage}
\begin{minipage}[t]{0.5\textwidth}
    \includegraphics[width=\textwidth]{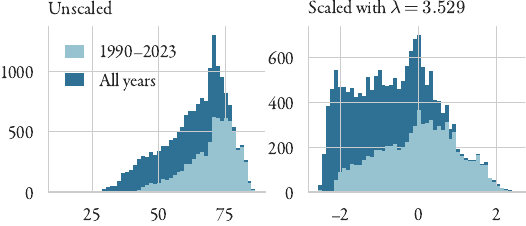}
\end{minipage}

\begin{minipage}[t]{0.5\textwidth}
    \cs{\textbf{E} Crude birth rate}
\end{minipage}
\begin{minipage}[t]{0.5\textwidth}
    \cs{\textbf{F} Crude death rate}
\end{minipage}
\hfill\vspace{-2mm}

\begin{minipage}[t]{0.5\textwidth}
    \includegraphics[width=\textwidth]{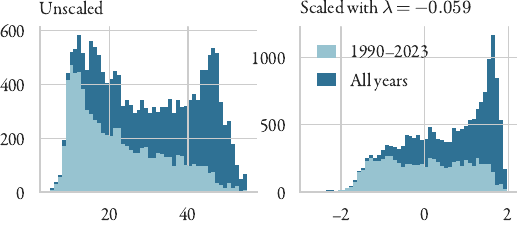}
\end{minipage}
\begin{minipage}[t]{0.5\textwidth}
    \includegraphics[width=\textwidth]{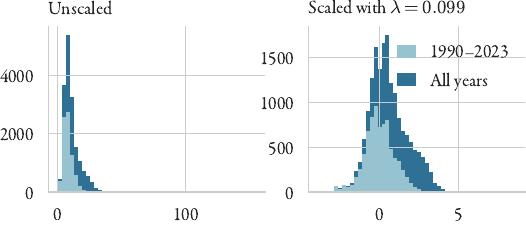}
\end{minipage}

\begin{minipage}[t]{0.5\textwidth}
    \cs{\textbf{G} Distance}
\end{minipage}
\begin{minipage}[t]{0.5\textwidth}
    \cs{\textbf{H} Migrant stock}
\end{minipage}
\hfill\vspace{-2mm}

\begin{minipage}[t]{0.5\textwidth}
    \includegraphics[width=\textwidth]{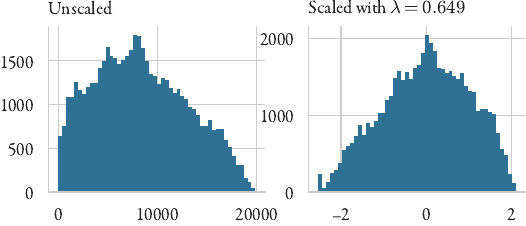}
\end{minipage}
\begin{minipage}[t]{0.5\textwidth}
    \includegraphics[width=\textwidth]{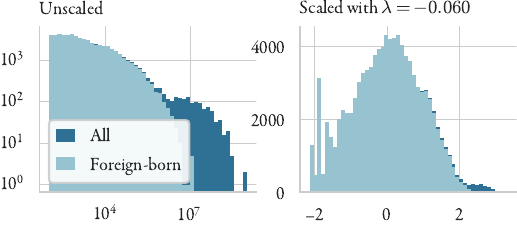}
\end{minipage}
\end{adjustbox}
\end{figure*}

\begin{figure*}[t!]
\begin{adjustbox}{minipage=\textwidth-20pt}
\begin{minipage}[t]{0.5\textwidth}
    \cs{\textbf{I} Refugee stock}
\end{minipage}
\begin{minipage}[t]{0.5\textwidth}
    \cs{\textbf{J} Change in refugee stock}
\end{minipage}
\hfill\vspace{-2mm}

\begin{minipage}[t]{0.5\textwidth}
    \includegraphics[width=\textwidth]{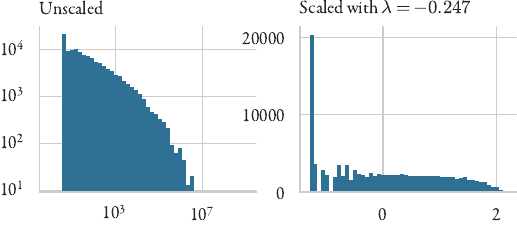}
\end{minipage}
\begin{minipage}[t]{0.5\textwidth}
    \includegraphics[width=\textwidth]{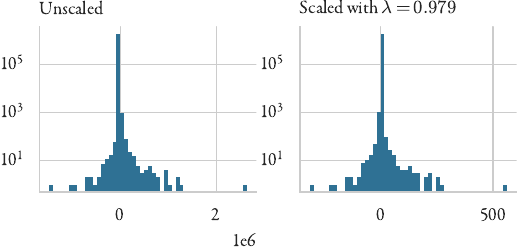}
\end{minipage}

\begin{minipage}[t]{0.5\textwidth}
    \cs{\textbf{K} Conflict-related deaths}
\end{minipage}
\begin{minipage}[t]{0.5\textwidth}
    \cs{\textbf{L} Bilateral trade}
\end{minipage}
\hfill\vspace{-2mm}

\begin{minipage}[t]{0.5\textwidth}
    \includegraphics[width=\textwidth]{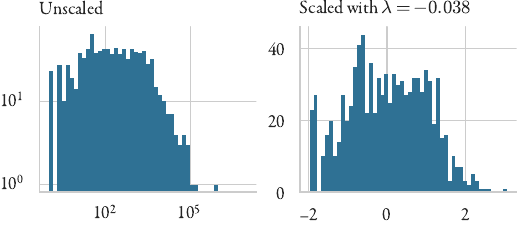}
\end{minipage}
\begin{minipage}[t]{0.5\textwidth}
    \includegraphics[width=\textwidth]{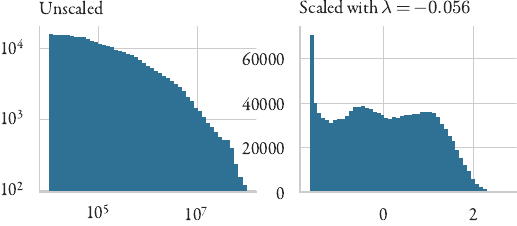}
\end{minipage}
\caption{Scaling the input covariates. The transformation parameter $\lambda$ is used to scale the raw data (left-hand figures) to an approximately normal distribution (right-hand figures) using the symmetrised Yeo-Johnson transform. The scaled values are then normalised to have zero mean and unit variance. The scaling $\lambda$ is determined via maximum likelihood optimisation for the period 1990--2023.}
\label{fig:SI:Covariate_scaling}
\end{adjustbox}
\end{figure*}

\subsection{Total population, Life Expectancy, Birth and Death rates}
We include the total population of the country of birth, country of origin, and country of destination for each edge $(i, j, k)$. These figures are taken from UN WPP dataset and scaled using a Yeo-Johnson transform (see fig. \ref{fig:SI:Covariate_scaling}{\cs{\textbf{C}}}--{\cs{\textbf{F}}}).
\subsection{Distance}
We use the \texttt{distw\_harmonic} distance covariate from the CEPII dataset as a measure of distance. Since values for some years are missing, we take the average over the period 1990--2023. Values for the British Virgin Islands and the Isle of Man are missing; we approximate these by using the values for the neighbouring US Virgin Islands, and inserting a value of 80km for the distance US Virgin Islands--British Virgin Islands. For the Isle of Man, we use distance values for the United Kingdom and insert a value of 300km for the distance United Kingdom--Isle of Man. The covariate is then scaled using a Yeo-Johnson transform, see fig. \ref{fig:SI:Covariate_scaling}{\cs{\textbf{G}}}.
\subsection{Linguistic similarity}
Linguistic similarity is taken the USITC Domestic and International Common Language Database \cite{USITC_DICL}. We use the \texttt{csl} (common spoken language) index. As detailed in the dataset description, values for 46 countries are missing. Following the authors, for these cases we use the average of the linguistic proximity \texttt{lpn} and common native language \texttt{cnl} instead.
\subsection{Religious similarity}
The Correlates of War database \cite{Correlates_of_war_religion} details the religious composition $\bm{\alpha}$ of each country, that is the share of adherents to major religions in each country. Missing values are taken from the CIA World Factbook \cite{CIA_Worldfactbook}. The religious similarity between two countries is then simply $\langle \bm{\alpha}_i, \bm{\alpha}_j \rangle$, where we exclude the `other' (\texttt{othrgenpct}) category. This gives a more accurate picture of religious proximity than the CEPII \texttt{comrelig} covariate, which only uses the share of Catholics, Protestants, and Muslims to construct a similarity score.
\subsection{EU Membership}
This is taken from the CEPII gravity dataset. We construct two covariates, $\text{EU}_{ij}$ and $\text{EU}_{jk}$, which are 1 if the indexed countries are both EU members, 0 else. As a binary covariate, it is left unscaled.
\subsection{Colonial ties}
Colonial relations between two countries are taken from the USITC gravity dataset (\texttt{colony\_of\_destination\_ever}). We include two covariates, $\text{COL}_{jk}$ and $\text{COL}_{ik}$, which are 1 if $j$ (or $i$) was ever a colony of the destination $k$, and 0 else. As a binary covariate, it is left unscaled.

\begin{figure}[b!]
\begin{adjustbox}{minipage=\textwidth-20pt}
    \includegraphics[width=\textwidth]{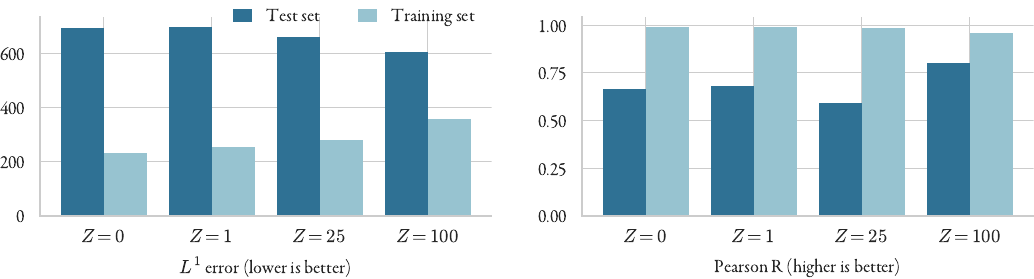}
\caption{Adjusting the dimension $Z$ of the latent state $\bm{z}$. $Z=0$ indicates a regular feed-forward architecture with no recurrent structure. The far left, the average $L^1$ error on the test and training set. Right: average correlation test and training set.}
\label{fig:SI:latent_space_validation}
\end{adjustbox}
\end{figure}

\begin{figure}[t!]
    \begin{adjustbox}{minipage=\textwidth-20pt}
\begin{minipage}{\textwidth}
    \cs{\textbf{A} Evolution of the $L^1$ prediction errors during training}
\end{minipage}
\hfill\vspace{-2mm}

\begin{minipage}{\textwidth}
    \includegraphics[width=\textwidth]{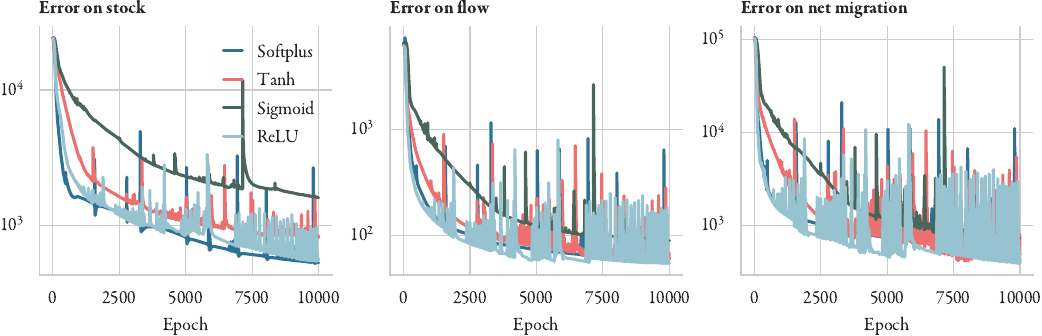}
\end{minipage}
\vspace{0.1em}

\begin{minipage}{\textwidth}
    \cs{\textbf{B} Relative error distribution of $\bm{\hat{T}}$}
\end{minipage}
\hfill\vspace{-2mm}

\begin{minipage}{\textwidth}
    \includegraphics[width=\textwidth]{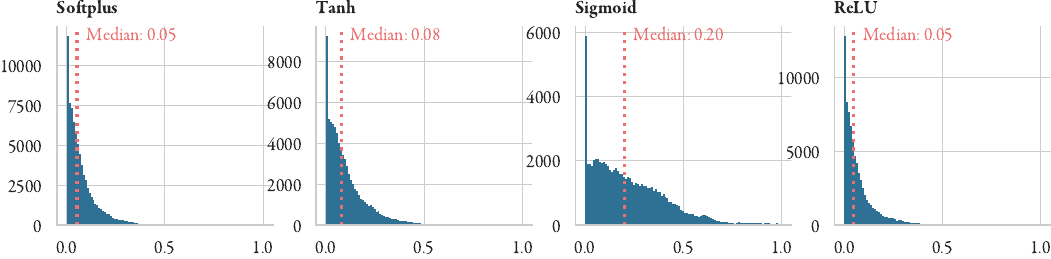}
\end{minipage}
\vspace{0.1em}

\begin{minipage}{\textwidth}
    \cs{\textbf{C} Examples of inferred flow corridors $T_{ijk}$}
\end{minipage}
\hfill\vspace{-2mm}

\begin{minipage}{\textwidth}
    \includegraphics[width=\textwidth]{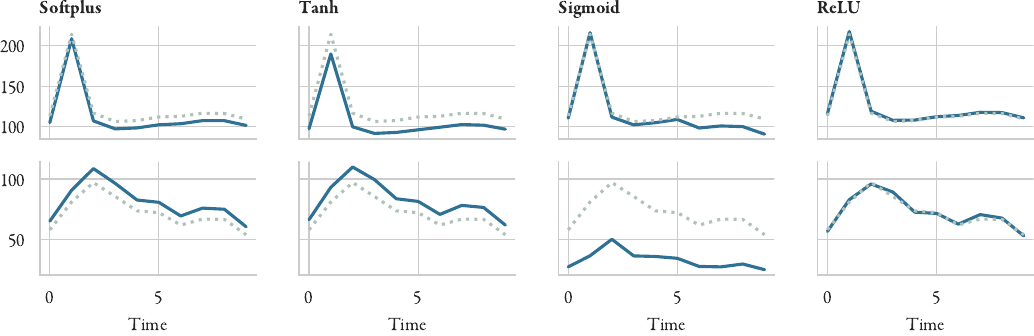}
\end{minipage}
\caption{Validation of the neural network approach on noiseless, fully observed synthetic data. In all cases, we use a 3-layer neural network with 20 neurons per layer, but vary the activation function on the hidden layer. \textbf{A} The evolution of the error on the stocks, flows, and net migration as the models train. All models converge to the true target data, though at different rates. \textbf{B} The distribution of the relative error on the full table $\bm{\hat{T}}$ after 10,000 epochs. The median error is indicated. \textbf{C} Predicted and true (dashed) values $T_{ijk}$ on two randomly selected edges. After 10,000 epochs (roughly 2 minutes of training on a GPU), the best-performing models have largely already converged to the true solution $\bm{T}$.}
\label{fig:SI:noiseless_validation}
\end{adjustbox}
\end{figure}

\begin{figure}[t]
    \begin{adjustbox}{minipage=\textwidth-20pt}
\begin{minipage}{\textwidth}
    \cs{\textbf{A} Evolution of the $L^1$ prediction errors during training}
\end{minipage}
\hfill\vspace{-2mm}

\begin{minipage}{\textwidth}
    \includegraphics[width=\textwidth]{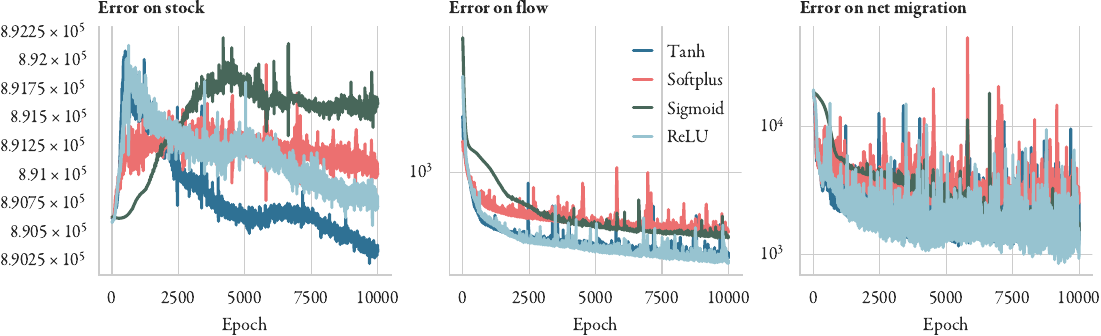}
\end{minipage}
\vspace{0.1em}

\begin{minipage}{\textwidth}
    \cs{\textbf{B} Architecture tuning}
\end{minipage}
\hfill\vspace{-2mm}

\begin{minipage}{\textwidth}
    \includegraphics[width=\textwidth]{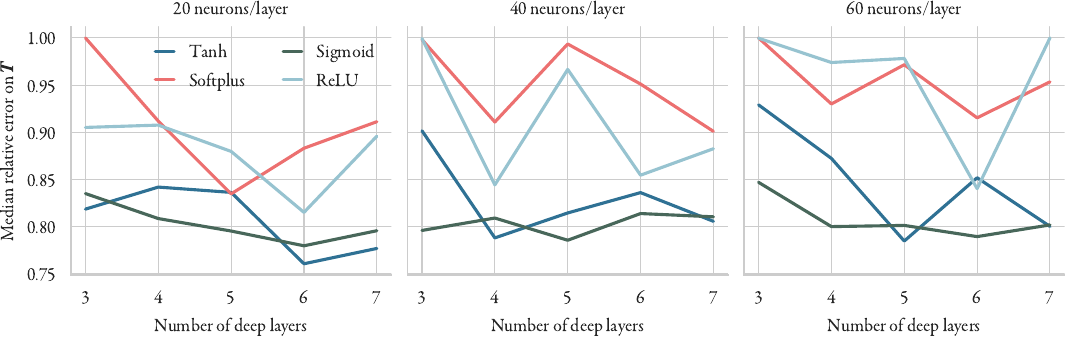}
\end{minipage}
\caption{Tuning of the neural network hyperparameters on noisy, partially observed synthetic data. \textbf{A} The evolution of the error on the stocks, flows, and net migration is now noticeably slower and more jagged than in the noiseless case. \textbf{B} Comparison of the median relative error on $\bm{\hat{T}}$ for various architectures.}
\label{fig:SI:noisy_validation}
\end{adjustbox}
\end{figure}

\section*{Model validation}
We validate our approach on a small synthetic dataset of migration flows between 30 destination and origin countries over 10 years. This is both to ensure our model is able to in principle infer the full flow table from observations of the stocks and flows, as well as to tune the hyperparameters. To do this, we select 30 random countries from the period 2010--2020, generate a random initial stock table $\bm{S}(t_0)$, and generate flows from
\begin{equation}
    T_{ijk} = \eta \exp(\langle \bm{\chi}_{ijk}, \bm{\alpha}\rangle),
\end{equation}
where $\bm{\chi}$ are the Yeo-Johnson-transformed covariate vectors, $\eta = 100$ a scalar, and $\bm{\alpha}$ a vector of random coefficients drawn from an i.i.d uniform distribution, $\alpha_i \sim \mathcal{U}[0, 0.5]$. Note that the vector $\bm{\chi}$ contains the stocks $S_{ij}$ and $S_{ik}$ themselves, as in the full model; these are calculated recursively from $\bm{T}$. This produces a full synthetic dataset of stocks, flows, and net migration vectors on which we can validate the approach. We first assume all flow, net migration, and stock values are observed, and train the neural network for 10,000 epochs. We use a 3-layer neural network with 20 neurons in the hidden layers, and the same CeLU activation function on the final layer as in the main manuscript. We vary the choice of activation function in the hidden layer. As shown in fig. \ref{fig:SI:noiseless_validation}, the full flow table $\bm{T}$ is then uniquely inferred from the data, though accuracy varies with the choice of activation function. Naturally, since we are observing the full flow table $\bm{F}$, the net migration data are fully determined by $\bm{F}$ and thus redundant. 

To generate a more realistic dataset, we repeat the above process, adding 10\% (multiplicative) noise to the stocks, 20\% noise to flows, and 5\% noise to the net migration data. We then randomly mask out 80\% of flow corridors (that is, entire origin-destination corridors), 80\% of net migration corridors, and 10\% of individual stock values (not corridors). To tune the hyperparameters of the model, we perform a grid search on the depth, width, and activation functions of the network, and compare prediction accuracy on the true flow table $\bm{T}$ (uncorrupted by noise) after 10,000 training epochs. Results are shown in fig. \ref{fig:SI:noisy_validation}, indicating that the hyperbolic tangent activation function, together with a high number of layers, gives best prediction results.

To select the dimension of the latent space, we select 20\% of flow corridors uniformly at random, and include only the remaining 80\% in the training data; then we train an ensemble of neural networks with different latent space dimensions $Z$ and compare performance. Fig. \ref{fig:SI:latent_space_validation} shows the $L^1$ error on the test set, the training loss, the average correlation on the test set, and the distribution of correlations on the flow corridors contained in the test set. As can be seen, a high latent space of dimension of 100 performs best of all the dimensions considered.

\begin{figure*}[t!]
    \begin{adjustbox}{minipage=\textwidth-20pt}
    \begin{minipage}{\textwidth}
        \cs{\textbf{A} Relative errors on the flow $\bm{F}$}
    \end{minipage}
    \hfill\vspace{-2mm}

\begin{minipage}{\textwidth}
    \includegraphics[width=\textwidth]{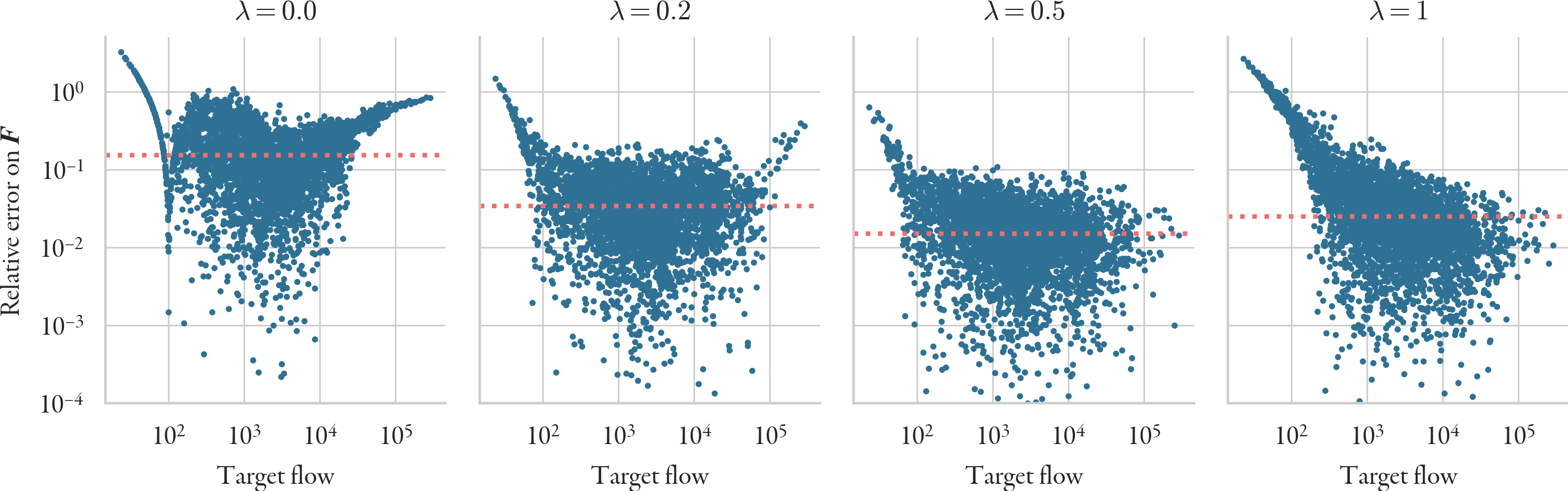}
\end{minipage}
\vspace{0.1em}

    \begin{minipage}{\textwidth}
        \cs{\textbf{B} Median relative error}
    \end{minipage}
    \hfill\vspace{-2mm}

\begin{minipage}{\textwidth}
    \includegraphics[width=\textwidth]{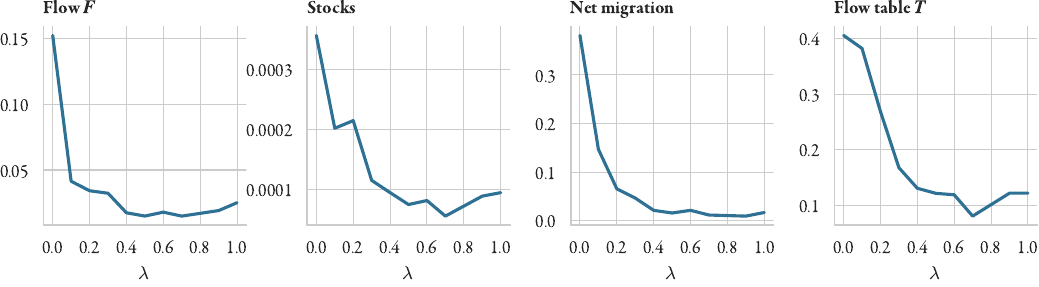}
\end{minipage}
\caption{Tuning the Yeo-Johnson scaling parameter $\lambda$. \textbf{A} A value of $\lambda=0.5$ balances the relative errors across different orders of magnitude of the target value. If $\lambda$ is too small, all values are penalised equally, leading to poor performance across all orders of magnitude. A large value of $\lambda$ leads to poor performance on small flows. The median relative error (red dotted line) is optimal for values around $\lambda=0.5$. \textbf{B} Median relative error on all four datasets as a function of $\lambda$.}
\label{fig:SI:Yeo_Johnson_scaling}
\end{adjustbox}
\end{figure*}

Finally, we set the Yeo-Johnson scaling parameters to a value of $\lambda=0.5$ for all target datasets, based on the following heuristic. The $0.5$ value, intuitively, lies `half-way' between the identity transform $\lambda=1$, and the logarithmic transform $\lambda=0.0$. In fig. \ref{fig:SI:Yeo_Johnson_scaling} we show the learned flow values on the noiseless datasets for different values of the transformation parameters, suggesting that values between 0.5 and 0.7 are optimal. Small values of $\lambda$ penalise all relative errors more equally, meaning a relative error of 50\% on a flow of 100 contributes more to the loss than an error of 10\% on a flow of 1 million---though in absolute terms these errors are of course orders of magnitude apart. For small $\lambda$, the model tries to optimise all targets equally well (or, in practice, equally badly). Large values of $\lambda$, conversely, mean that small flows are disregarded, and only the large values optimised. Centre values around $\lambda=0.5$ balances these considerations. \vfill 

\newpage 

\vspace*{4in}

\newpage
\section*{Additional figures}
\begin{figure}[t!]
\begin{adjustbox}{minipage=\textwidth-20pt}
    \centering
    \includegraphics[width=\textwidth]{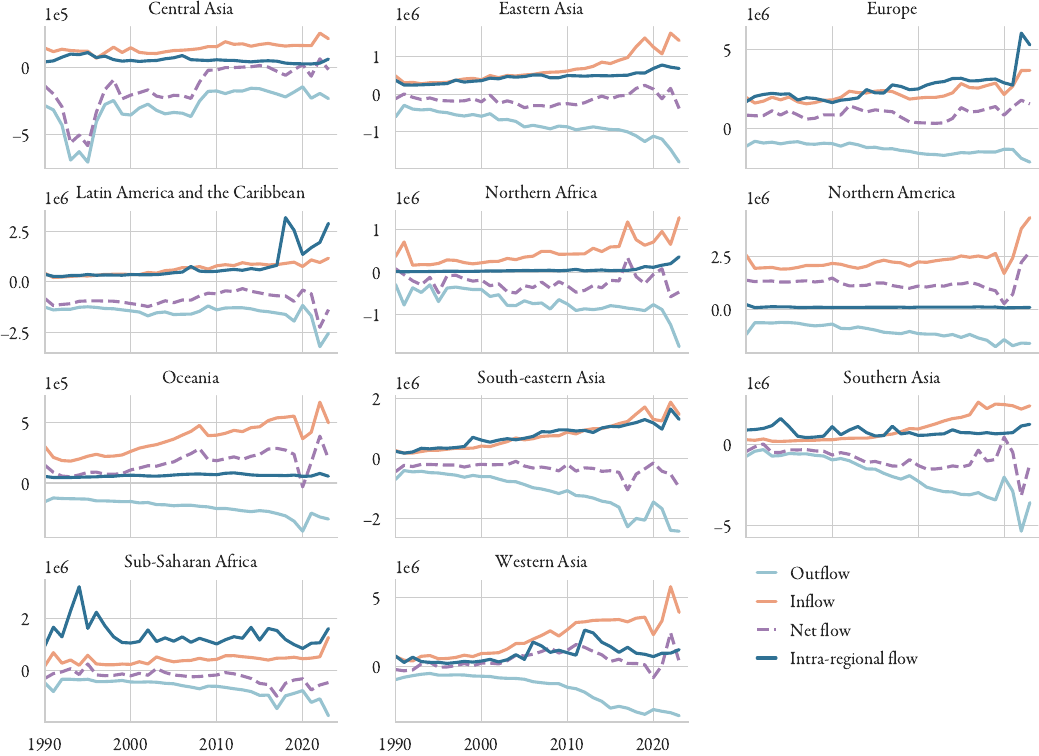}
    \caption{The same plots as in figure \ref{fig:regional_flows} for all geographic regions.}
    \label{fig:SI:regional_flows}
\end{adjustbox}
\end{figure}

\begin{figure}[t!]
\begin{adjustbox}{minipage=\textwidth-20pt}
    \centering
    \includegraphics[width=\textwidth]{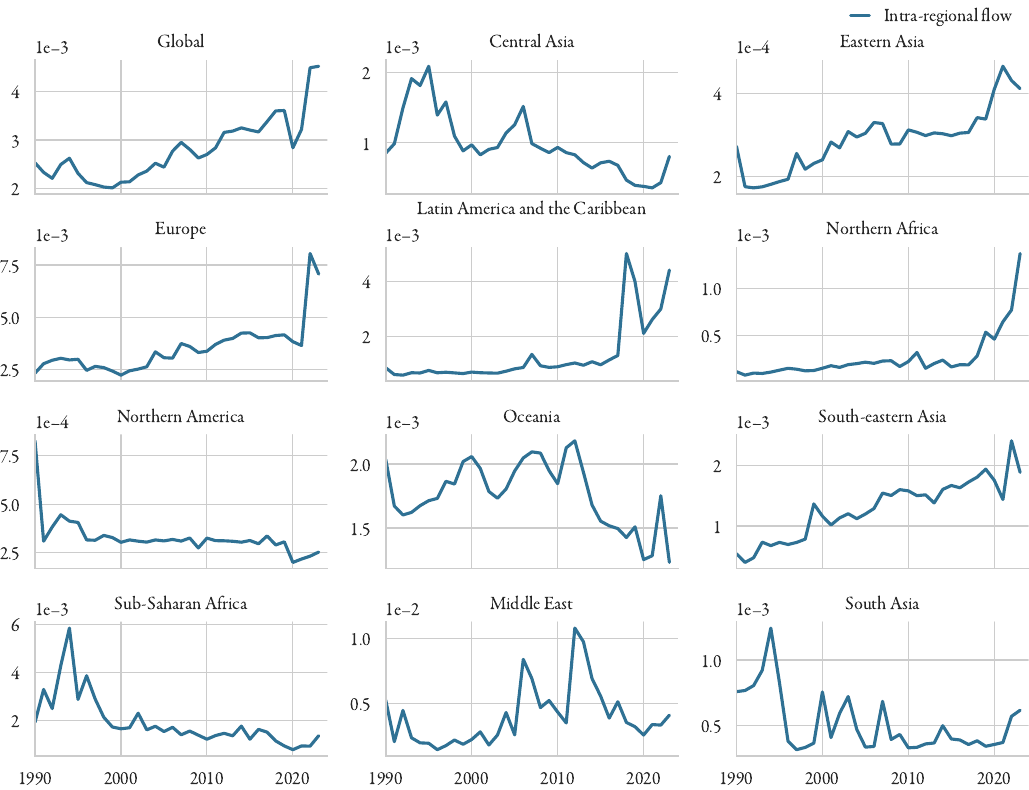}
    \caption{Intra-regional per capita flows for all regions. Also shown are the total global flows (top left).}
    \label{fig:SI:regional_flows_per_cap}
\end{adjustbox}
\end{figure}

Here we show some additional plots for the regional flows (fig. \ref{fig:regional_flows}) in the main manuscript. Figure \ref{fig:SI:regional_flows} shows the outflow, inflow, net flow, and intra-regional flow for all regions; figure \ref{fig:SI:regional_flows_per_cap} shows the intra-regional flows per capita. Also shown is the per capita total global migration flow: as is visible, the same trend of rising global migration emerges, even when accounting for the global population increase.

\vspace*{4in}

\newpage

\bibliographystyle{bibstyle}
\bibliography{library}

\end{document}